\documentclass{article}

\usepackage{arxiv}

\usepackage[utf8]{inputenc} 
\usepackage[T1]{fontenc}    
\usepackage{url}            
\usepackage{booktabs}       
\usepackage{amsfonts}       
\usepackage{nicefrac}       
\usepackage{microtype}      
\usepackage{graphicx}
\usepackage{doi}
\usepackage{amsmath}
\usepackage{subfloat}
\usepackage{subfig}

\title{A Deep Learning Object Detection Method for an Efficient Clusters Initialization}

\author{
Raphaël Couturier\\
Univ. Bourgogne Franche-Comté (UBFC),\\ FEMTO-ST Institute,\\ France\\
   \And
  Hassan N. Noura  \\
Univ. Bourgogne Franche-Comté (UBFC),\\ FEMTO-ST Institute,\\ France\\

   \And
   Ola Salman\\
   American University of Beirut,\\
   Electrical and Computer Engineering Department,\\ Beirut 1107 2020, Lebanon\\
  \And
  Abderrahmane Sider\\
  Laboratoire LIMED, University of Bejaia, Algeria
}

\date{ }

\begin{document}

\maketitle

\begin{abstract}
Clustering is an unsupervised machine learning method grouping data samples into clusters of similar objects. In practice, clustering has been used in numerous applications such as banking customers profiling, document retrieval, image segmentation, and e-commerce recommendation engines. However, the existing clustering techniques present significant limitations, from which is the dependability of their stability on the initialization parameters (e.g. number of clusters, centroids).  Different solutions were presented in the literature to overcome this limitation (i.e. internal and external validation metrics). However, these solutions require high computational complexity and memory consumption, especially when dealing with big data. In this paper, we apply the recent object detection Deep Learning (DL) model, named YOLO-v5, to detect the initial clustering parameters such as the number of clusters with their sizes and centroids. Mainly, the proposed solution consists of adding a DL-based initialization phase making the clustering algorithms free of initialization. Two model solutions are provided in this work, one for isolated clusters and the other one for overlapping clusters. The features of the incoming dataset determine which model to use. Moreover, The results show that the proposed solution can provide near-optimal clusters initialization parameters with low computational and resources overhead compared to existing solutions. 
\end{abstract}


\section{Introduction}


Clustering is an efficient solution to split a data-set into a set of clusters, that are characterized by the high similarity within the same cluster and the high distance between different clusters~\cite{xu2015comprehensive,christophe}. Mainly, the clustering methods are unsupervised machine learning methods that can be parametric (probability-based) or non-parametric~\cite{sinaga2020unsupervised}. The non-parametric clustering methods are often based on an empirical function that measures the similarity or dissimilarity between the data points~\cite{fahad2014survey}. On the other hand, the parametric methods assume some mapping functions or probability distributions and try to find their best parameters to optimally cluster the data points. In addition, the clustering methods can be classified into two main types, which are hierarchical and partitional. With partitional clustering methods, which are most employed, the data point belongs to one cluster~\cite{barioni2014open}. However, with hierarchical clustering, there are nested clusters, such that a data point can belong to many clusters at a time. 

\subsection{Problem Formulation}
Unfortunately, the clustering algorithms suffer from several challenges, preventing their reliability and efficiency. A main challenge is that the stability and convergence of the clustering algorithms depend of the initialization parameters. For some clustering algorithms, the number of clusters must be given a priori, as in the case of k-means. In fact, the clusters number is mostly unknown, given that clustering is generally applied on unlabeled data-sets. To determine the number of clusters, several metrics can be employed, from which we include the well-used metrics in~\tablename~\ref{exisitingMetrics}. However, using these metrics is hindered by several limitations, especially in terms of computation and memory overhead, when dealing with large-scale data-sets. Therefore, there is a need for an efficient clustering initialization solution, which is the main contribution of this paper.\\


\subsection{Motivation and Contributions}
Proposing a clustering initialization solution that requires low computation overhead is crucial since the existing initialization techniques are the most computation-consuming functions, especially for frequently updated and/or large data-sets.  In this paper, an efficient solution for clustering parameters initialization is presented. 
The proposed solution consists of training a supervised deep neural network (i.e. Yolo-v5) in order to predict the required initialization parameters of existing clustering algorithms. To the best of our knowledge, we are the first to apply object detection algorithms to find appropriate clusters initialization parameters. The proposed solution can predict the number of clusters and their corresponding centroids, in addition to the size of each cluster. Moreover, two models  are presented in this work, one for isolated clusters and the other for overlapping clusters. The properties of the input data set determine which model to use.\\

To assess our approach, we present extensive experimental results on several clustering algorithms, i.e. k-means , Fuzzy C Means (FCM), with and without pre-computed initialization parameters using the proposed approach. The results show an improvement in both performance and stability of many of the considered clustering algorithms. \\

Finally, it is worth noting that our approach achieves high generalization since our training is done on data with small number of centroids, while at the testing phase we are able to detect the centroids of data presenting higher number of clusters.  Furthermore, this paper presents a good application of transfer learning to solve issues of existing unsupervised machine learning algorithms.

\subsection{Organization}
The rest of this paper is organized as follows: In Section~\ref{sec:back}, we list and briefly describe some clustering algorithms, the internal validation metrics used for clusters initialization, in addition to the used object detection DL-based approach. Then, Section~\ref{sec:prop} presents the proposed detection clustering initialisation parameters. In Section~\ref{sec:perf}, experimental results with real data-sets are presented to validate the efficiency of the proposed solution. Finally, the paper is concluded in Section~\ref{sec:conc}.


\section{Background and Preliminaries}
\label{sec:back}
This section presents a set of well-known clustering algorithms such as k-means and FCM. For each of them, we present their required initialization parameters. Then, the advantages and limitations of existing well-known clustering algorithms are listed. Afterwards, several internal metrics used for initial cluster numbers detection are presented and described. Let us indicate that the proposed solution is also internal since we consider that we do not have any information about data labels. Finally, the employed YOLO-v5 DL-based object detection is described. Let us indicate that YOLO-v5 is used as proof of concept in the proposed solution and any object detection model can be used if it can ensure high detection accuracy. \\


In the following, we will use these notations : $X = \{x_1,\; \ldots ,\;x_n\}$ is a set of $n$ data points in a $d$-Euclidean space of dimension $R^d$. These data points can be clustered in $m$ different clusters of centroids $C = \{c_1,\;\ldots, \;c_m\}$.

\subsection{Clustering Algorithms}
Clustering algorithms have been extensively explored in the literature and implemented in a set of substantive areas~\cite{xu2015comprehensive,alhawarat2018revisiting,meng2018new}. Before dwelling into more details, let us recall that all the considered algorithms in this work are based on the Expectation-Maximisation (EM) algorithm~\cite{dempster1977maximum}. The EM algorithm is an iterative algorithm, where each iteration consists in two steps: computing the expectation (E-Step) and maximizing it (M-Step). This probabilistic technique is generally used to solve Maximum Likelihood problems. It transforms an optimization problem (minimizing an objective function) into a probabilistic problem solved through this simple yet very powerful technique as is the case with k-means, and FCM.\\

\subsubsection{K-means}
\label{sec:back:km}
In this part, the k-means algorithm, which is one of the most famous and ancient unsupervised learning algorithms~ \cite{mcqueen1967some}, is described.  

The k-means algorithm consists of the following steps :
\begin{enumerate}
    \item Set the number of clusters (centers) $m$
    \item Initialize centroids by shuffling the data-points and then randomly selecting $m$ centers without replacement
    \item Iterate until the maximum number of allowed iterations is reached or the difference between two successive sets of centers is under some error threshold\\
    \begin{enumerate}
    \item Updating the membership matrix: for each data-point, compute the euclidean distance to each old centroid, find the minimum distance for each data-point, assign that data-point to the centroid with which it has minimum distance (E step)
    \item Updating the centroids: find the new centroids by computing the average of each newly determined cluster according to the updated membership matrix (M Step)
    \end{enumerate}
\end{enumerate}

Each data point $x_i$, $i=1,\;2,\;\ldots, \;n$ is considered as a member of only one cluster $k$ of center $c_k$, with $k=1,\;2,\;\ldots, \;m$, if it has a minimal euclidean distance to $c_k$ compared to other cluster centers.
The k-means algorithm minimizes the inertia, which is defined as the sum of the distances between data-points and their centroids, as can be seen in Eq.~\ref{eq:J:KM}.\\ 

The big issue with k-means is that the number of clusters $m$ should be given a priori, but usually, this number is unknown in real applications, where the data is unlabeled. In addition, another limitation of k-means is that its stability is affected by the initialization parameters because the initial centroids are usually randomly chosen from the data-set in the so-called LLoyd's implementation. A more sophisticated approach, called "k-means++", uses a smarter heuristic for setting the initial number of centroids to achieve faster convergence, but the number of clusters should in all cases be known.\\

The recent x-means algorithm~\cite{pellegx}, a modified version of k-means, does not necessitate the a priori knowledge of the number of clusters, but instead, it relies on the Bayesian Information Criterion (BIC) measure to define the number of clusters. However, the stability of x-means is still in function of the initialization parameters as k-means. In addition, calculating the BIC measure presents a high computation overhead, especially with large data-sets. By contrast, our proposed solution, based on the object detection concept, detects efficiently the clusters number, possible centroids, and clusters sizes. Let us note that k-means minimizes the objective function defined by Eq.~\ref{eq:J:KM}, which is simply the sum of the distances from each data point to its corresponding centroid.\\

\begin{equation}
J_{m}(X,c)=\sum_{i=1}^{n}\sum_{k=1}^{m}\lVert x_{i} - c_{k}\rVert^{2}
\label{eq:J:KM}
 \end{equation}

\subsubsection{Fuzzy C Means}
\label{sec:back:FCM}
With k-means, each data-point may be a member of at most one cluster. While this might be well-adapted for many applications, in some cases, a data-point might belong to several clusters. The FCM algorithm, first presented by Dunn~\cite{dunn1973fuzzy} and later improved by Bezdek~\cite{bezed1981pattern}, leverages the fuzzy algebra to express the simultaneous membership of a data-point to different classes (clusters). The FCM algorithm computes what is called a soft partition of the data-set in contrast to k-means which computes a hard data partition. In order to compute a hard partition with FCM, it is sufficient to consider the maximum membership degree of a data-point as its unique cluster. The FCM algorithm tries to minimize the objective function given in Eq.~\ref{eq:J:FCM}, where $z$ is called the fuzziness parameter initialized to a value between $2$ and $3$, and $\lVert .\rVert^{2}_{A}$ stands for any mathematical distance.
 
\begin{equation}
J_{z}(X,c)=\sum_{i=1}^{n}\sum_{k=1}^{m} (u_{ki})^{z} \lVert x_{i} - c_{k}\rVert^{2}_{A}
\label{eq:J:FCM}
 \end{equation} 
The FCM algorithm consists of the following steps : 
 \begin{enumerate}
    \item Randomly initialize the membership matrix $U$ such that $\sum_{k=1}^{m} u_{ki} = 1$ for any data-point $x_i$
    \item Update the centroids using the membership matrix $U$ (E-Step)
    \item Update the membership matrix $U$ using the last computed centroids in the previous step (M-Step)
    \item Stop if $\lVert U^{t+1} - U_{t}\rVert^{2}_{A} < \epsilon$ or maximum number of allowed iterations is reached, otherwise make another iteration of steps $2$ and $3$.
\end{enumerate}

The FCM algorithm needs only the set of initial centroids but it must be fed with a proper membership matrix. A workaround may consist of running one extra "M-Step", after feeding the algorithm with a set of centroids. This helps to compute a proper membership matrix, that is consistent with the input centroids. Just like k-means, the FCM algorithm stability is very dependent on the initial membership matrix, which is randomly chosen. Our approach promise is to handle this limitation very efficiently since the initial $U$ matrix will be tightly linked to the detected centroids. \\

\tablename~\ref{exisitingClustersIssues} presents a set of well-known clustering algorithms with their advantages and disadvantages. Their main issue, as indicated in this table, is that they require information about the number of clusters or the clusters sizes. \\

\begin{table*}[!ht]
\caption{Overview of the advantages and disadvantages of commonly used clustering algorithms}
\begin{tabular}{|p{2cm}|p{7cm}|p{7cm}|}
\hline
\textbf{Clustering
Algorithm} &\textbf{Advantages} & \textbf{Disadvantages} \\ \hline
\textbf{k-means} & \begin{itemize} \item This algorithm is simple and has a linear relative computational complexity. \end{itemize}
 & \begin{itemize}
 \item The user must specify the number of clusters to be used.
\item With a cluster of irregular maps, output is poor performance.
 \end{itemize}
 \\\hline
\textbf{Mean Shift} & \begin{itemize} \item The number of clusters or classes is not required. \end{itemize} &\begin{itemize} \item The user must specify the window size. \end{itemize}\\\hline
\textbf{DBSCAN} & \begin{itemize} \item The number of clusters or classes is not required.
 \item It can detect outliers and irregularly shaped clusters.\end{itemize}& \begin{itemize} \item The user must specify the window size.
\item For different clusters densities, the performance is reduced. \end{itemize}\\\hline
\textbf{Expectation-Maximization}&  \begin{itemize} \item  It is capable of detecting clusters with ellipsoidal forms.\item Each point is given a membership probability. \end{itemize}& \begin{itemize} \item The user must specify the number of clusters beforehand.\end{itemize}\\\hline
\end{tabular}
\label{exisitingClustersIssues}
\end{table*}

To fix this challenge, several metrics (internal or external: with labels) that can be used to detect cluster numbers are presented in \tablename~\ref{exisitingMetrics}. However, these methods suffer to deal with big data-set and they are described in the next.

 \begin{table*}[!ht]
\caption{Existing metrics to detect clusters number for the k-means clustering algorithm}
\begin{tabular}{|p{1cm}|p{4cm}|p{11cm}|}
\hline
Reference & Metric & Description \\ \hline
 \cite{kass1995bayes}  & Bayesian Information Criterion (BIC) & 
BIC is a criterion for measuring and selecting models. It relies on the principles of Bayesian inference and probability. The model complexity is penalized by BIC, so more complex models would have a lower score and therefore be less likely to be chosen.  \\\hline
 \cite{bozdogan1987model}  & Akaike Information Criterion (AIC) &  It is suitable for models that fit into the maximum likelihood estimation system, like BIC. The lower are the AIC and BIC, the better is the clustering performance. \\\hline
 \cite{dunn1973fuzzy}  & Dunn’s index (DN) &  DN defines sets of clusters that are compact, with a very small variation between cluster members, and large separation between clusters. The higher is the value of Dunn's index, the better is the clustering performance. The optimal amount of clusters is the number of clusters that maximises the Dunn's index.  \\\hline
 \cite{davies1979cluster}  & Davies-Bouldin index (DB) & It calculates the average similarity between each cluster and its most similar one. The DB validity index aims to maximize the distances between clusters while minimizing distances between the cluster centroid and its data objects. \\\hline
 \cite{rousseeuw1987silhouettes}  & Silhouette  Width  (SW) & It is a statistic that measures how similar an object is to its own cluster versus other clusters. The silhouette value ranges from -1 to 1. A high silhouette value is well suited to its own cluster but poorly related to neighboring clusters. Positive and negative high silhouette widths (SW) indicate the objects that are correctly clustered and those that are incorrectly clustered, respectively. It is well known that objects with a SW validity index of zero or less are difficult to be clustered. \\\hline
 \cite{calinski1974dendrite}  & Calinski  and  Harabasz  index  (CH)  & This metric is the ratio of the sum of between-cluster dispersion and inter-cluster dispersion for all clusters. It is known also as the Variance Ratio Criterion. The higher is this score, the better is the clustering performance.  \\\hline
 \cite{tibshirani2001estimating}  & Gap  statistic  &  It is a statistical hypothesis test-based cluster validity measure. At each value of the cluster number, the gap statistic compares the variation in within-cluster dispersion to that predicted under an appropriate reference null distribution. The smallest is the number of clusters, the best is the clustering performance. \\\hline
\end{tabular}
\label{exisitingMetrics}
\end{table*}

\subsection{Validation Metrics}
\label{sec:back:val}
In this part, we present the state-of-the-art methods for clusters number estimation. To find the appropriate number of clusters $k$, several validity metrics were presented and they can be divided into two main classes~\cite{rendon2011internal}: external and internal.\\
\begin{enumerate}
    \item External metrics (data labels) are employed to evaluate a clustering method performance by comparing the obtained cluster memberships with their initial class labels.
    \item Internal metrics are used to evaluate the correctness of the obtained clusters structure by focusing on the intrinsic information of the data itself. 
\end{enumerate}
  
In general, the most advanced methods of estimating the cluster numbers can be described as follows.: 
\begin{enumerate}
    \item For $k = 1,\; 2, \ldots,\; k_{max}$, recognize $k$ clusters in the specified data-set using the clustering algorithm.
    \item Apply the cluster number estimation method to all of the output clusters, for $k = 1,\; 2, \ldots,\; k_{max}$.  
    \item To estimate the number of clusters, use the appropriate selection criteria (max, min, etc.).
\end{enumerate}





\subsection{YOLO Deep Learning Object Detection Model}
\label{sec:back:YOLOv5}

Object detection is among the classical computer vision problems to identify which objects are in the image and their corresponding locations. The object detection issue is more complex than the classification problem that consists of recognizing objects but without indicating their locations in the image. Moreover, images containing more than one object cannot be classified.\\

In this paper, we use YOLO-v5, a recent update of the YOLO family. YOLO was the first object recognition model to merge bounding box estimation and object identification in one end-to-end differentiable network. Darknet is the environment under which YOLO is written and maintained. In comparison to previous YOLO models, YOLO-v5 is the first YOLO model developed with the PyTorch framework, and it is more lightweight and simple to use compared to previous YOLO variants. \\

YOLO-v5 is based on a smart Convolutional Neural Network (CNN) for real-time object detection. This algorithm divides the image into regions and calculates the bounding boxes and probabilities for each region. The predicted probabilities are used to weight these bounding boxes. The algorithm needs only one forward propagation pass through the neural network to make predictions, so it "only looks once" at the image. It then outputs known objects together with the bounding boxes after a non-max suppression (which ensures that the object detection algorithm only recognizes each object once).\\


\section{Proposed Model}
\label{sec:prop}
\begin{figure*}[!htbp]
\centering
\includegraphics[scale=0.5]{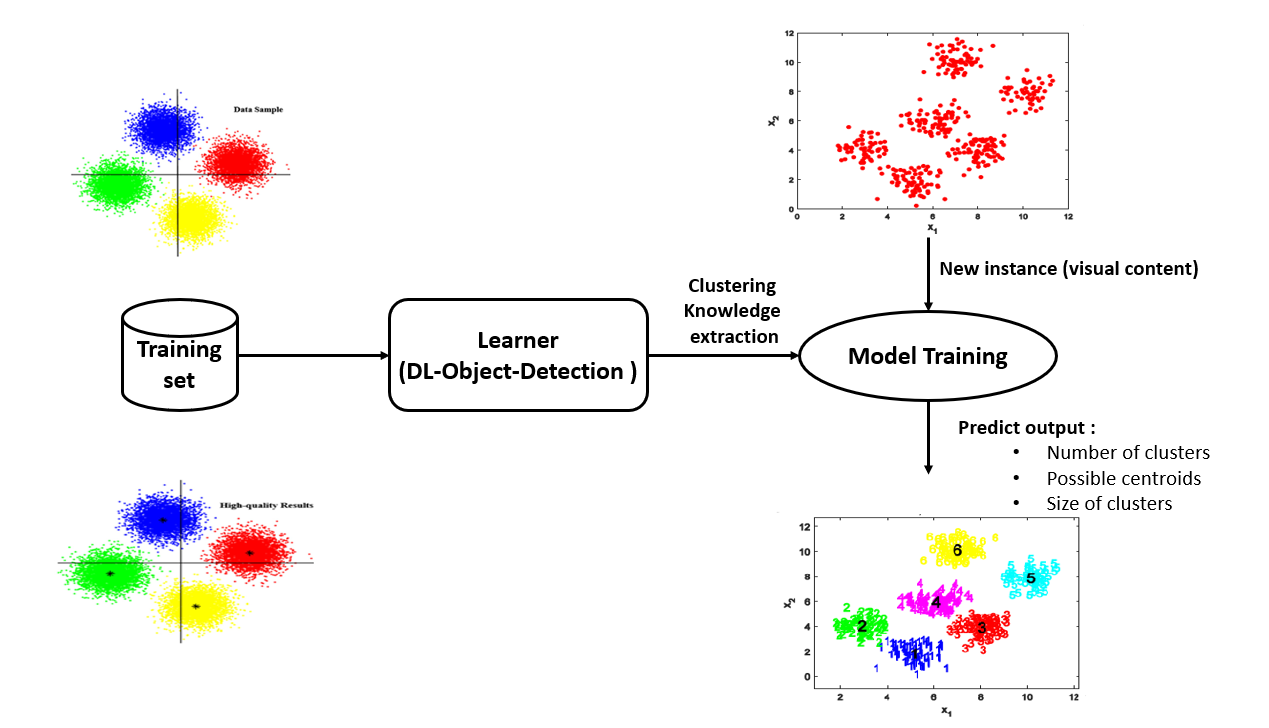}
\caption{The proposed solution consists of training the YOLO-V5 model to detect clusters initialization parameters such as the number of clusters and centroids }
\label{fig:proposedvar}%
\end{figure*}

The structure of the proposed solution, illustrated in \figurename~\ref{fig:proposedvar}, uses YOLO-v5 as a learner to detect clusters (as objects) with their centroids and sizes. The YOLO-v5 DL-based object detection method is used in this paper to construct a model that can make the clustering algorithm free of initialization. This model finds automatically the appropriate clustering initialization parameters such as the number of clusters, possible centroids and clusters sizes. In the following, we present the implementation details of the proposed solution. 

\subsection{Model Description}
The input of the proposed model is a transformed data in 2D feature space. The proposed solution assumes that a high-dimensional data-set can be represented in a lower dimension space, where an inter-cluster pairwise distance is preserved among clusters. Given a transformed data-set of $n$ points $X = { x_1,\ldots, x_n }$, where $x_i \in R^2$, the clustering algorithms, such as k-means, partition the data-set into $k$ clusters $C_1, C_2,\ldots,C_k$, where each cluster $C_j$ is represented by a cluster center $c_j$. Let $k$ denotes the correct number of clusters in a data-set. To build accurate clustering models that can achieve high clustering accuracy, $k$ and other clustering parameters (e.g. centroids and clusters sizes) are required as input. Getting a good estimate of these parameters and especially $k$ is not a straightforward task. Several metrics were presented previously to determine the number of clusters. However, computing these metrics requires a high computation and resources overhead.\\

The input of the proposed solution is an image representing the clustered data. Thus, with large data-set, the computation overhead of the proposed solution will remain the same, contrarily to the existing techniques that require an exponential computation complexity in function of the data size, as shown in~\figurename~\ref{fig:ClusterstimeIdentification}. 


\subsection{Data Description}
\begin{figure*}[!htbp]
\centering
\includegraphics[scale=0.5]{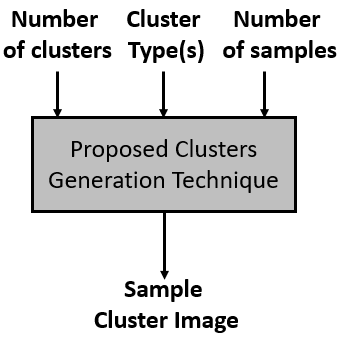}
\caption{Proposed clusters generation method}
\label{fig:generatedata-set}%
\end{figure*}

The YOLO-v5 model was trained using a generated data-set. This data-set consists of a large set of images generated with random configuration, as illustrated in \figurename~\ref{fig:generatedata-set}.  Each image contains a number of clusters (between 2 and 12) of random size, having each a number of samples between 20000 and 50000 points. Moreover, the form of clusters can be of one of the following types: Gaussian Mixture Model, noisy moon, blob, no structure, etc. \\

The obtained training images have a size equal to 640 $\times$ 640. These images are processed by scaling them between $[0;\;1]$. The testing images consist also of a set of generated clusters with random configuration. The importance of the proposed solution is that it is flexible (can add new cluster forms) and takes into account different conditions (dense or not). Let us indicate that the generated training/testing images are pre-processed to obtain the required labels files that are considered as input to YOLO-v5. \\

In total, the data-set contains 1000 images. Let us indicate that the generated clusters with random configurations can have equal or different number of points. Also, the obtained clusters can be well-separated or very close, in addition to having equal or variable variances. \\

\begin{figure*}[!htbp]
\centering
\subfloat[][$nc$=2]{\includegraphics[scale=0.125]{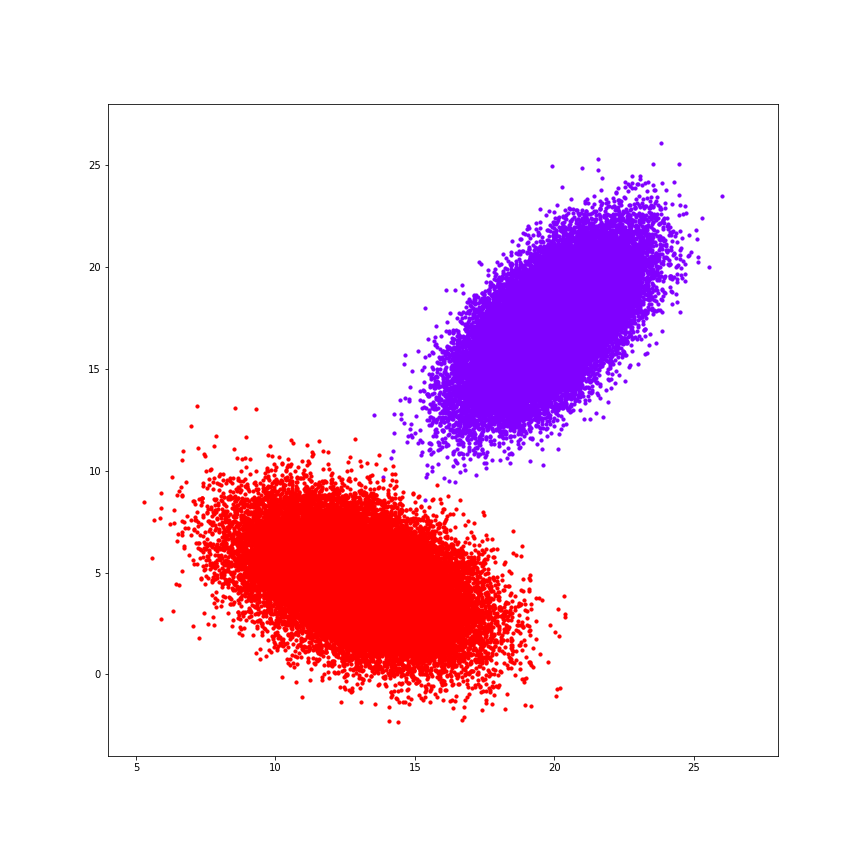}}
\subfloat[][$nc$=4]{\includegraphics[scale=0.125]{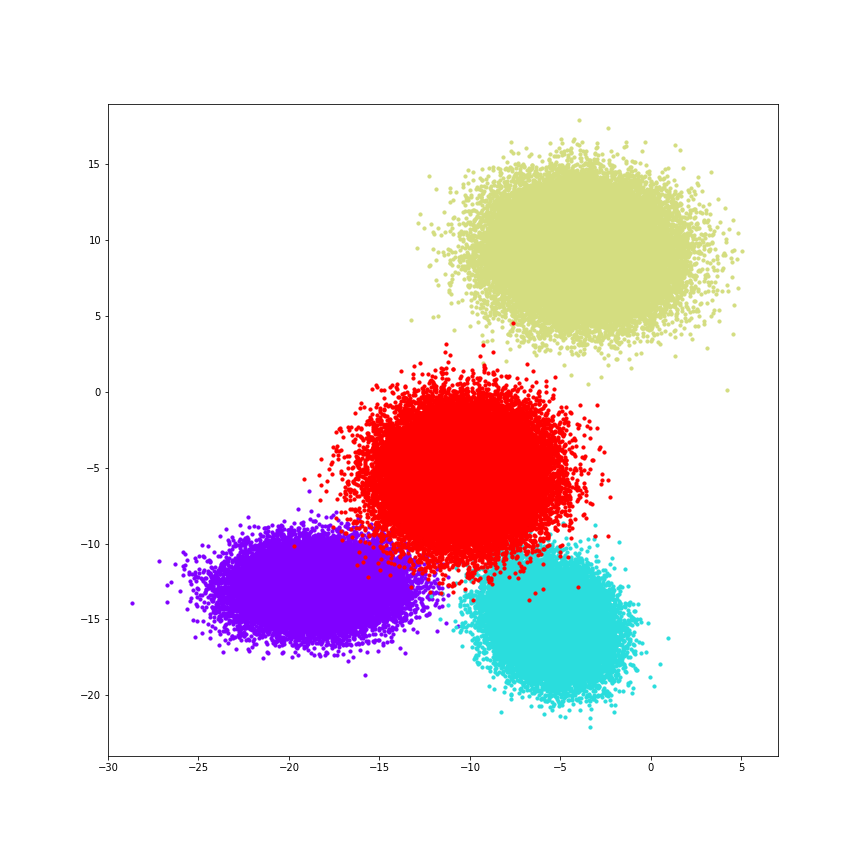}}
\subfloat[][$nc$=8]{\includegraphics[scale=0.125]{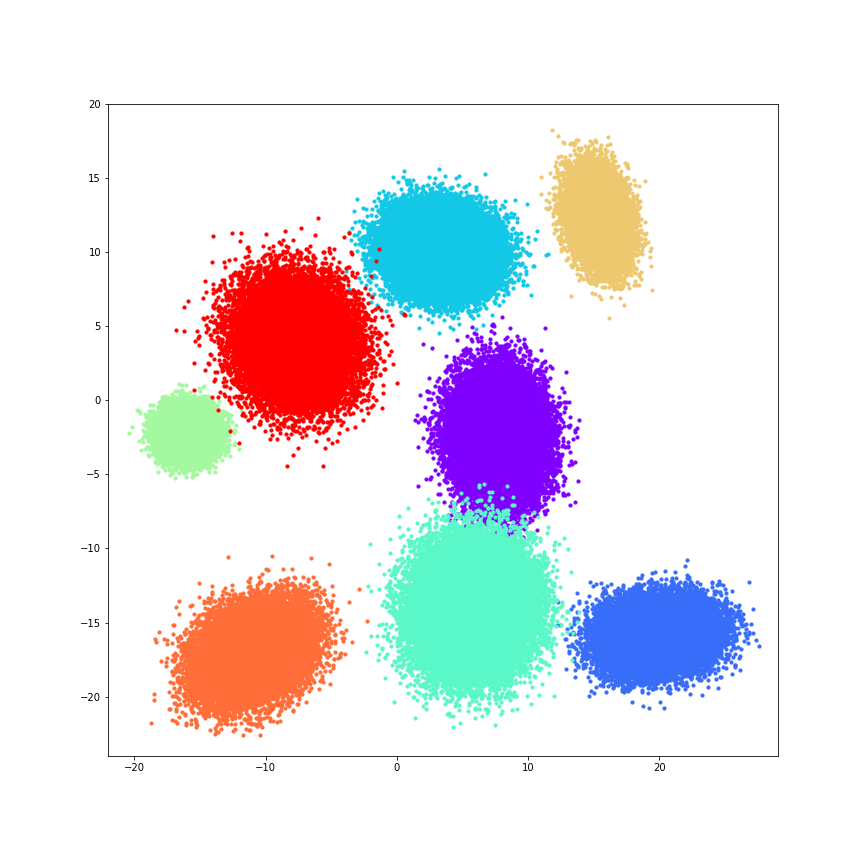}}
\subfloat[][$nc$=10]{\includegraphics[scale=0.125]{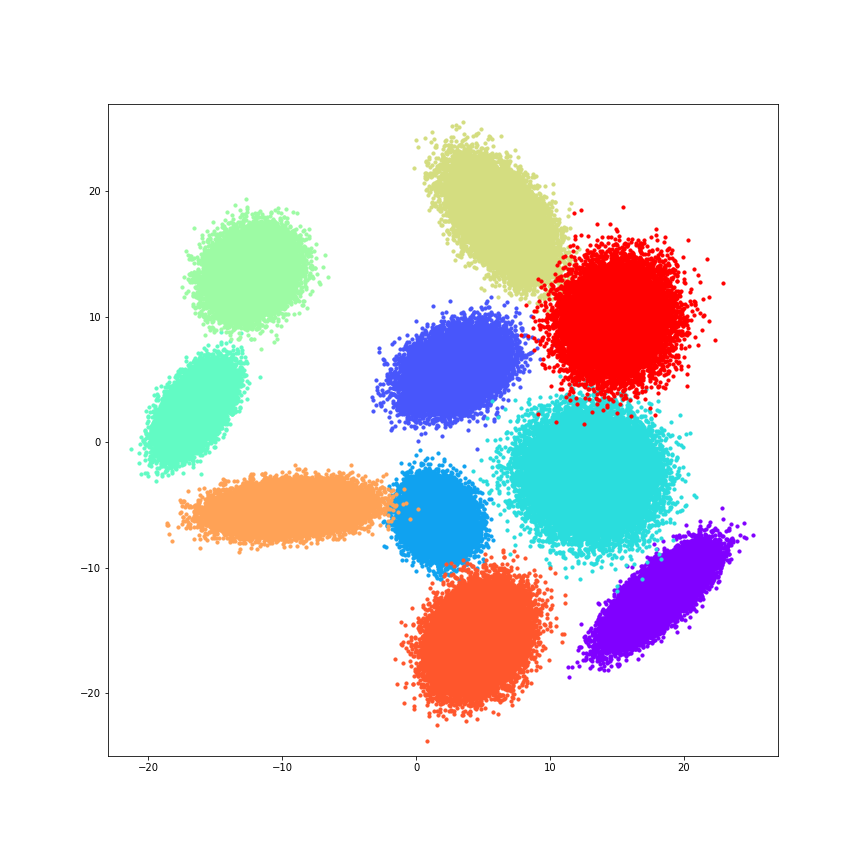}}\\
\caption{Examples of generated data-sets used to train YOLO-v5 object detection model}%
\label{fig:data-setoriginal}%
\end{figure*}

In \figurename~\ref{fig:data-setoriginal}, we present several examples of training images generated from the 2-variate with variable mixture components (2 dimensions with different clusters number). 

\subsection{Model Implementation}
The YOLO-v5 model implementation of~\cite{YOLOv5PyTorch} was used to build our proposed learner model. To train this model, the SGD optimizer was used with an initial value of $10^{-2}$ and a batch of size 16. All the other parameters are the standard parameters of the Yolo-V5 code.\\


In the training phase, the Yolo-V5 model learns to recognize all the clusters with 20 epochs. At the inference phase, once the model detects the bounding boxes, the center for each cluster is computed and this value will be the initial value for the initial cluster. As Yolo-V5 being very efficient and lightweight, it can detect quickly objects (clusters) and can be implemented on GPU or CPU (after being trained).

\section{Performance Analysis}
\label{sec:perf}
In this section, we present several performance and robustness tests that were applied on 100 testing iterations (images) to demonstrate the efficiency and robustness of the proposed solution. We have trained the object detection model (supervised learning) to obtain the initial clustering parameters such as the clusters number $k$, centroids and clusters sizes. \\

Thus, we will first compare our obtained results with the results obtained when applying the listed metrics in Section~\ref{sec:back:val} to validate the robustness and efficiency of the proposed solution. Furthermore, the proposed solution was applied with different clustering algorithms: k-means~\cite{likas2003global}, x-means~\cite{pelleg2000x}, and FCM~\cite{maji2007rfcm} with or without initialization parameters. For each test iteration, the clusters image consists of a number of clusters varying between 2 to 12, with random positions, directions, and sizes. We choose the random configuration of the clusters generation to check if that affects the performance of the proposed method. Furthermore, all clusters in the training and testing phases have random number of points, variance and also clusters separation distance. 

\subsection{Testing Methodology: Notations and Settings}
In this subsection, the testing methodology and the various parameters that might affect the effectiveness of the proposed approach are presented. \\ 

 When testing the proposed method, two possible cases should be considered:

\begin{enumerate}
\item The case of overlapping clusters data, where the data-set contains overlapping clusters.
\item The case of of separated clusters data, where the data-set contains non-overlapping or separated clusters.
\end{enumerate}

overlapping separated clusters
Therefore, two different trained models should be built, and the selection of which model to be used depends on the input data-set. Ideally, data clustering aims to satisfy two conditions, which are:
 \begin{itemize}
    \item Compactness: where the cluster around each center is dense.
    \item Separation: where different clusters are far apart.
\end{itemize}

When these conditions are met, the first trained model can be used and a set of experimental results that validate the effectiveness and the robustness of this model will be presented in the new section (see \figurename~\ref{fig:Clustersprocess}). \\

\begin{figure*}[!htbp]
\centering
\subfloat[][Detected $nc$=5, BIC=6, AIC=7]{\includegraphics[scale=0.12]{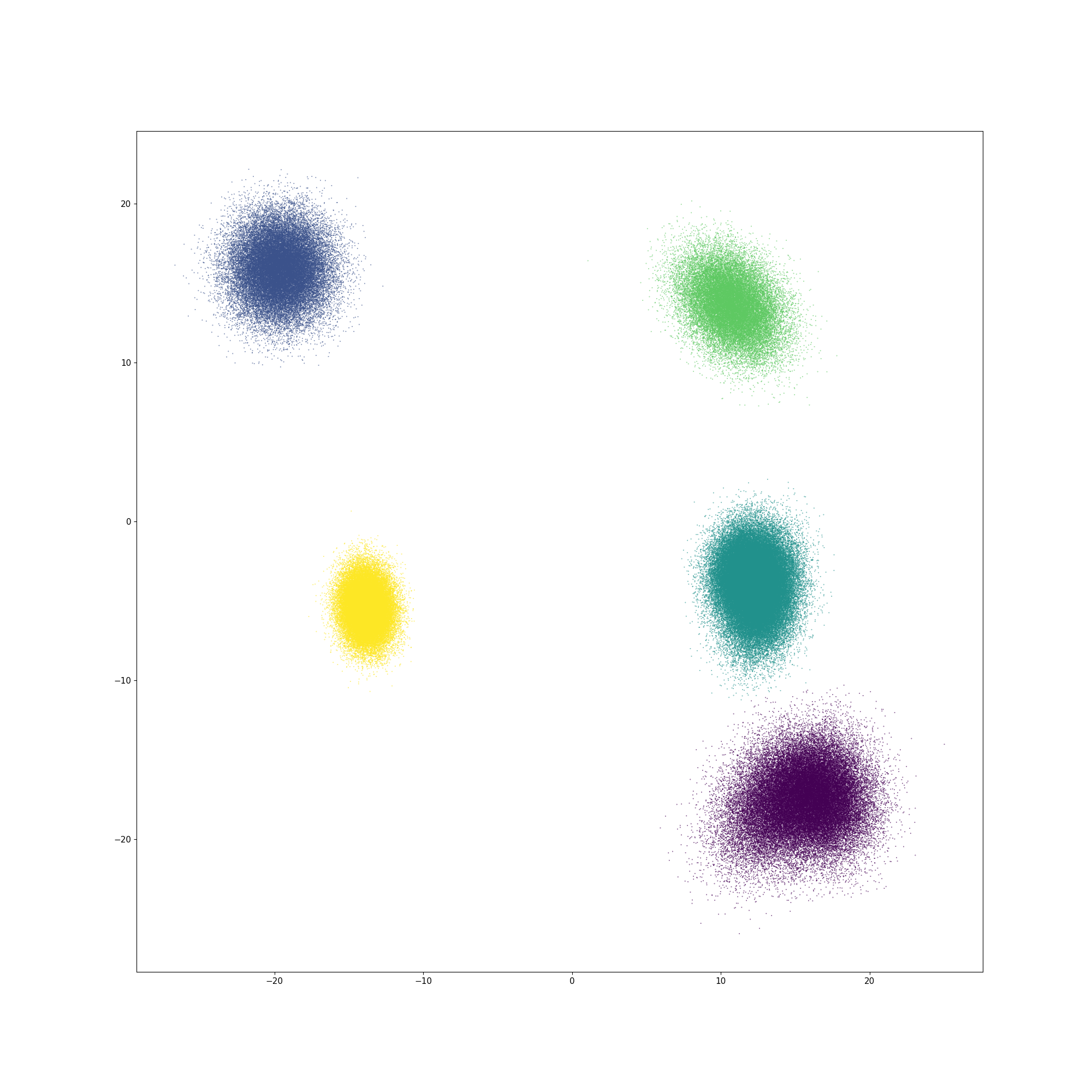}}
\subfloat[][Detected $nc$=4, BIC=5,	AIC=5]{\includegraphics[scale=0.12]{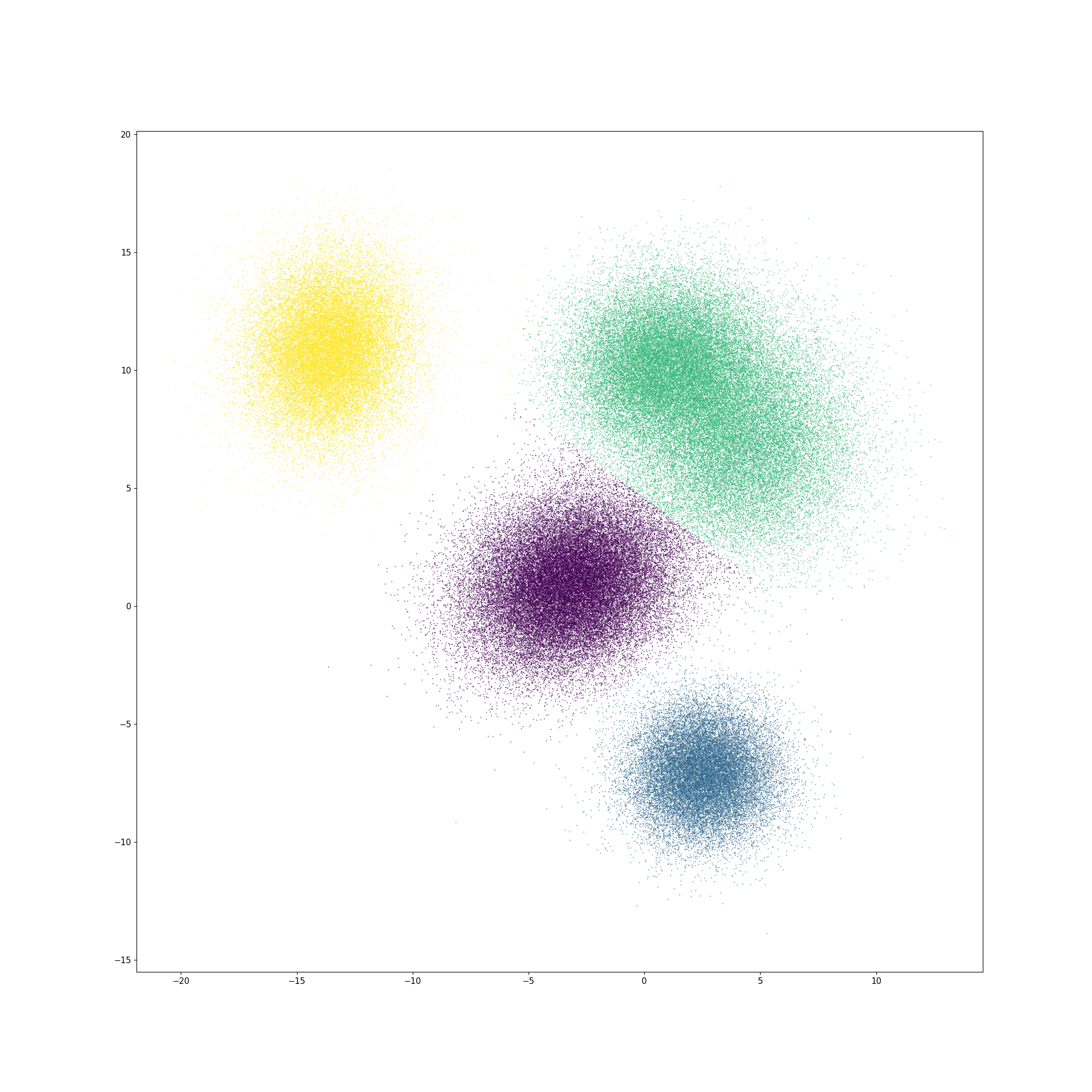}}
\subfloat[][Detected $nc$=3, BIC=4,	AIC=4]{\includegraphics[scale=0.12]{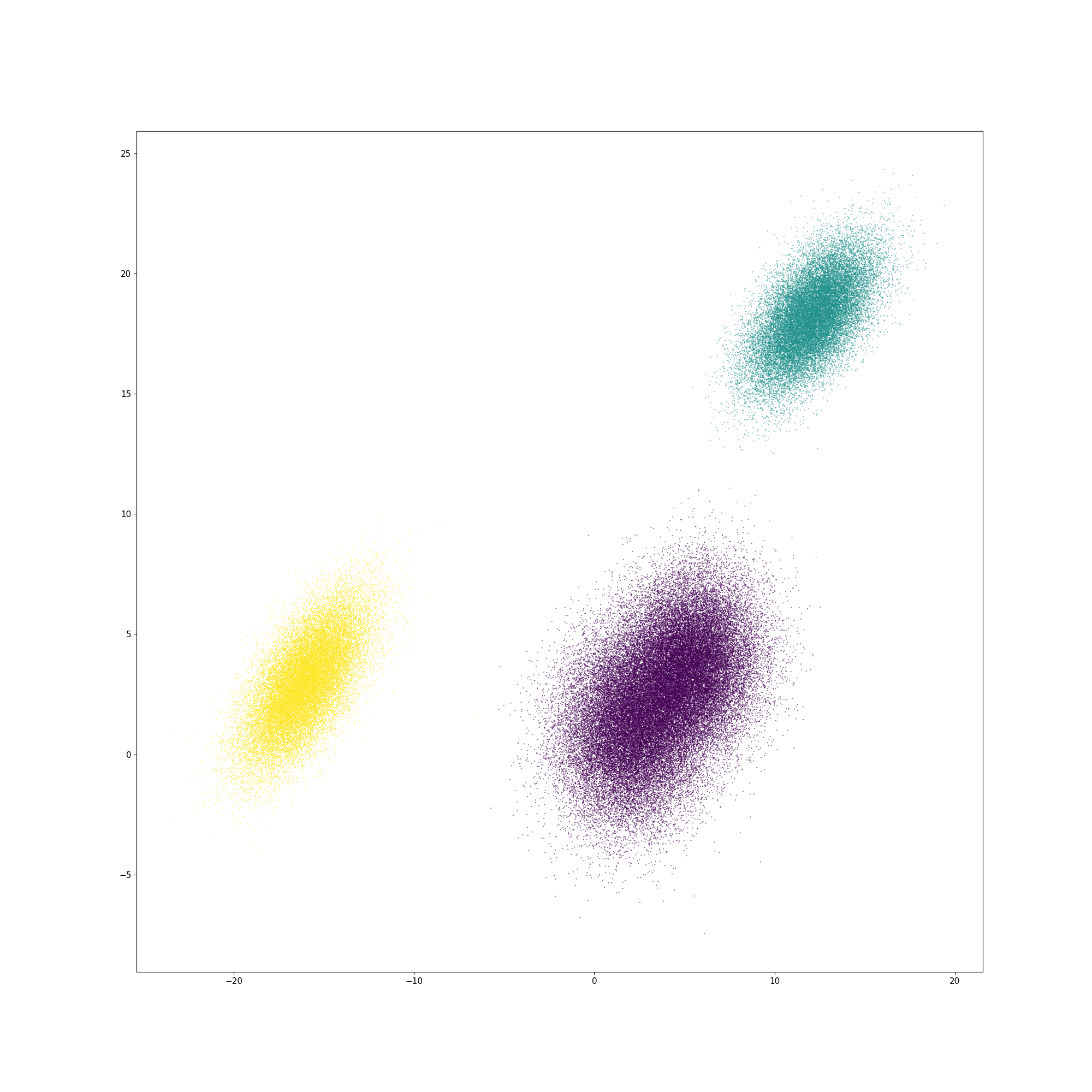}}\\
\subfloat[][Detected $nc$=4, BIC=6,	AIC=8]{\includegraphics[scale=0.12]{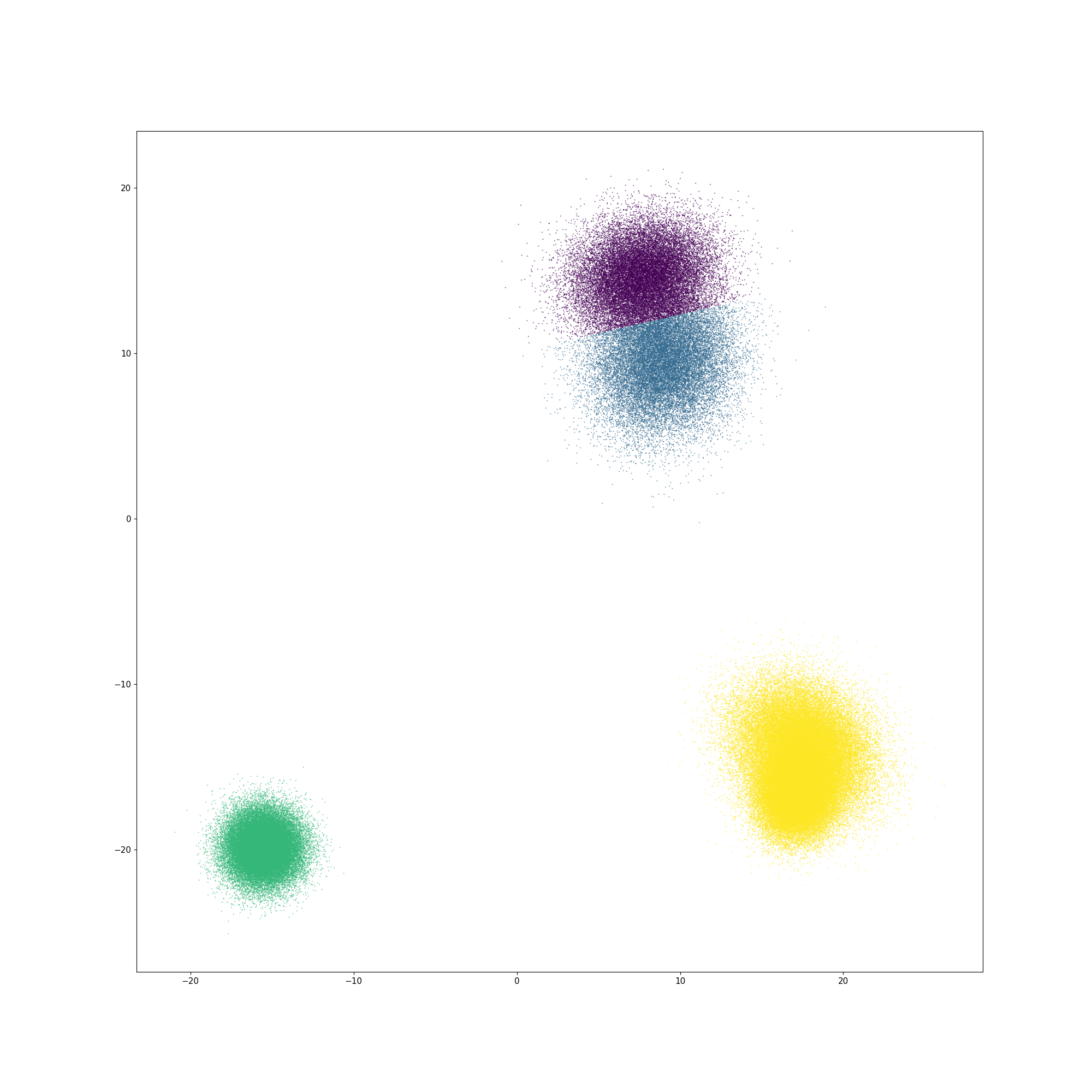}}
\subfloat[][Detected $nc$=3, BIC=4	AIC=4]{\includegraphics[scale=0.12]{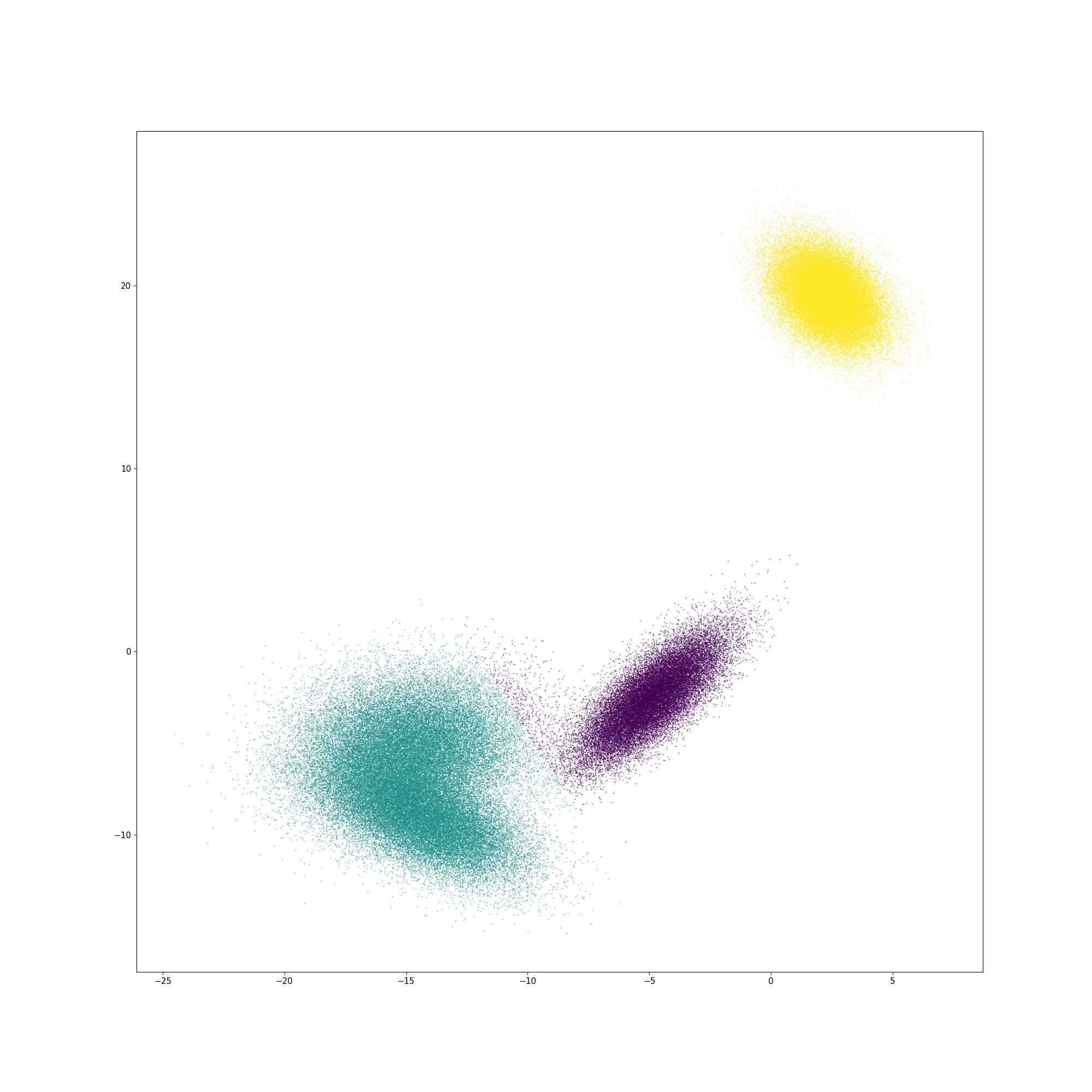}}
\subfloat[][Detected $nc$=7, BIC=8,	AIC=8]{\includegraphics[scale=0.12]{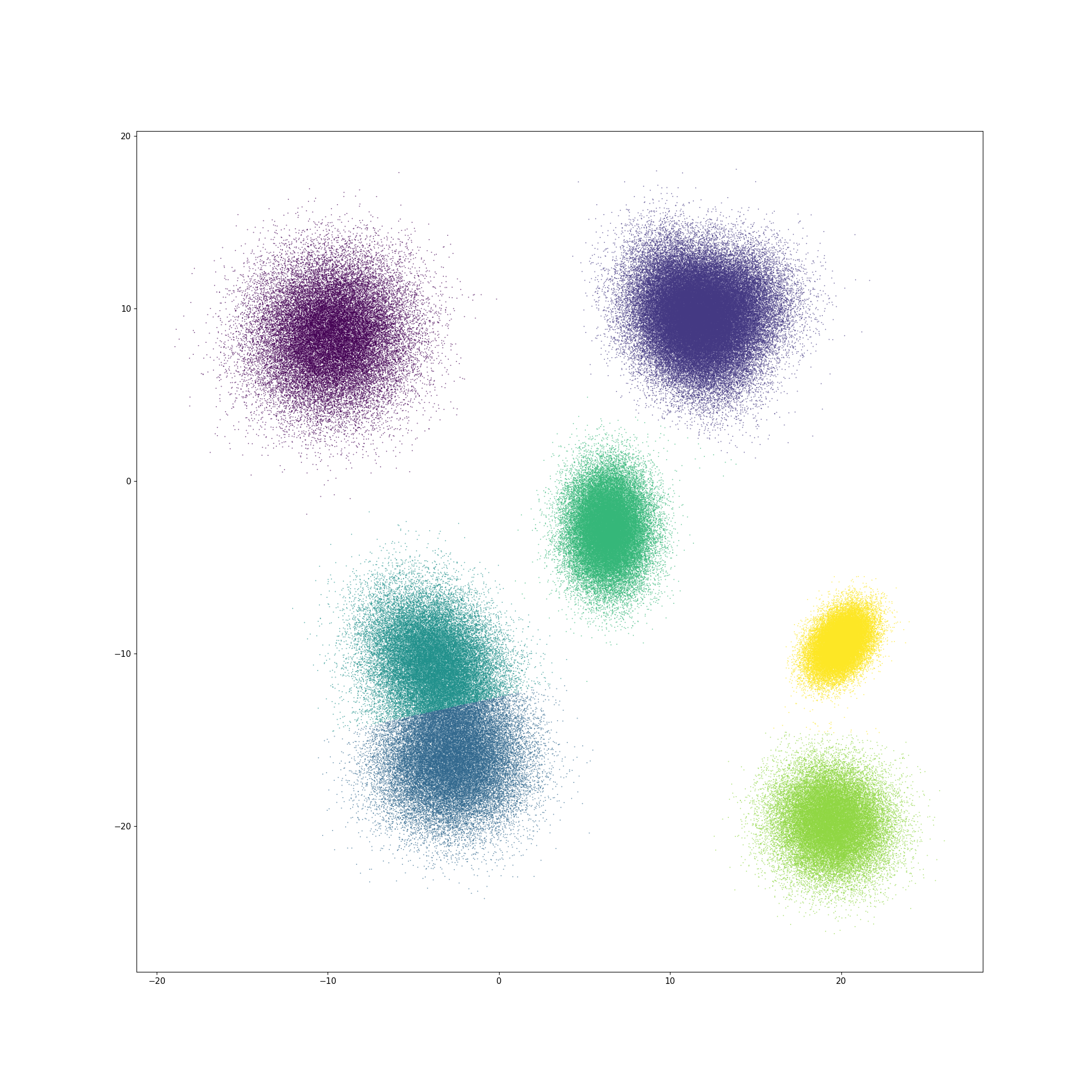}}\\
\subfloat[][Detected $nc$=4, BIC=5,	AIC=5]{\includegraphics[scale=0.12]{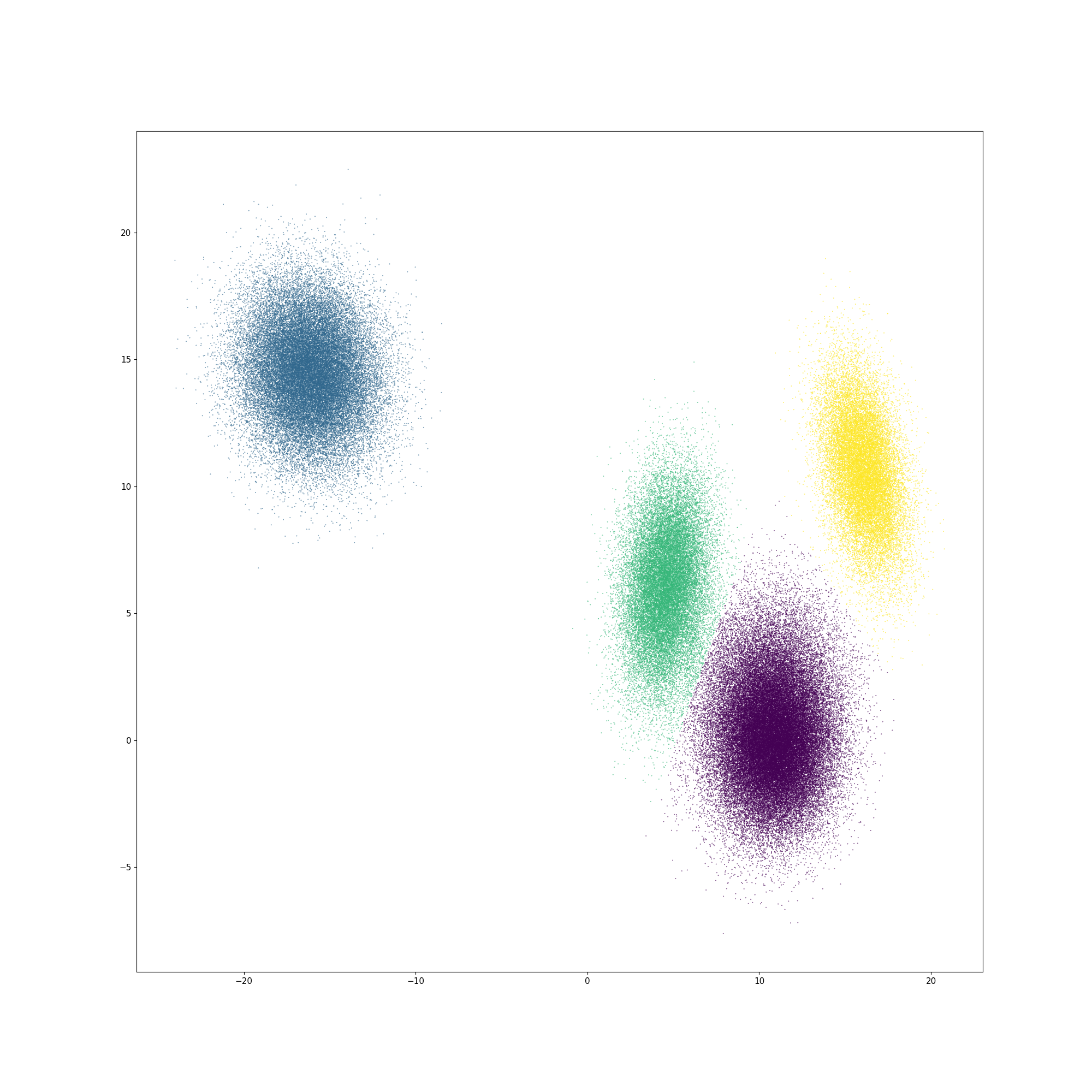}}
\subfloat[][Detected $nc$=5, BIC=8,	AIC=8]{\includegraphics[scale=0.12]{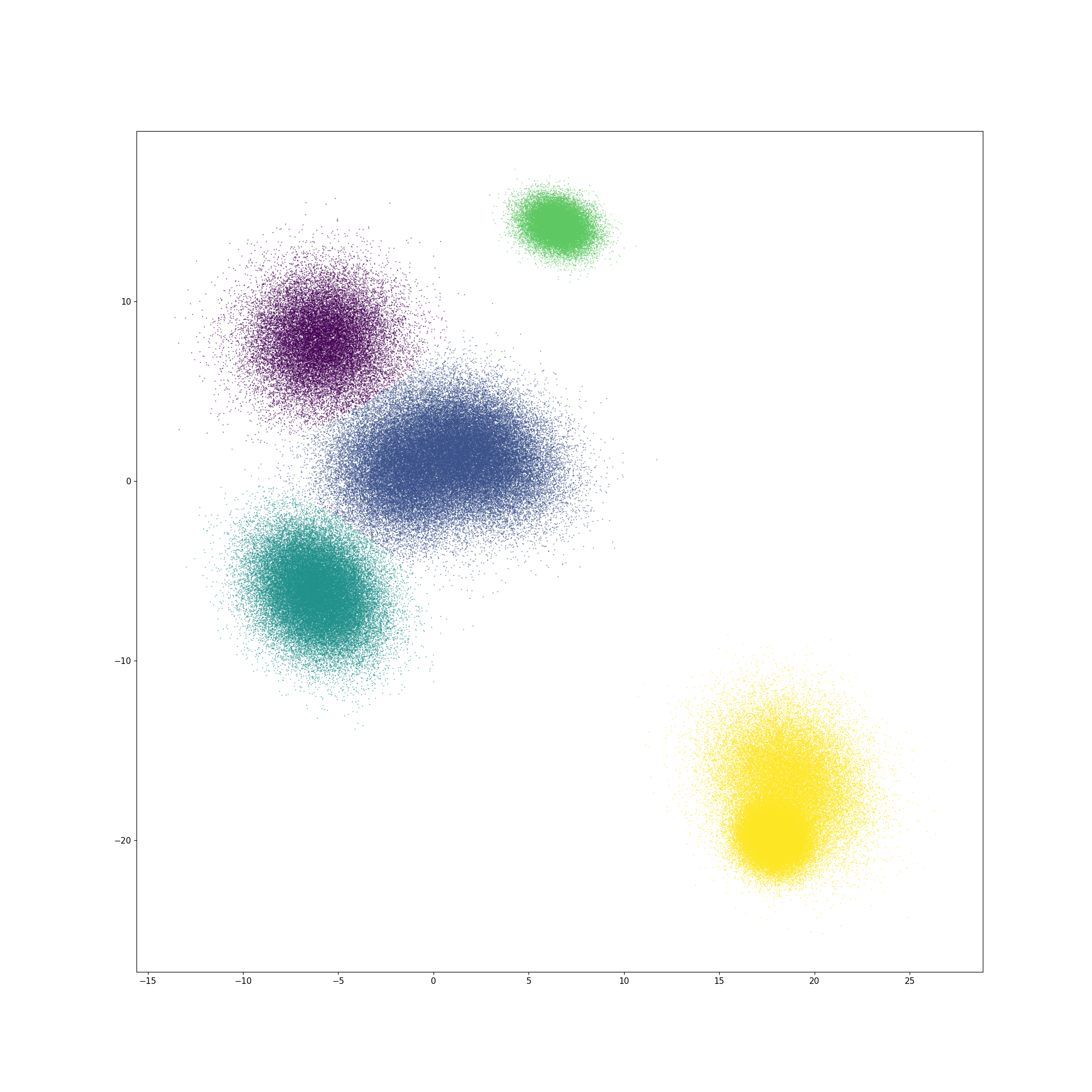}}
\caption{Results of the overlapping model for which the number of centroids detected by the proposed approach is different compared to other classical approaches (AIC, and BIC)  }
\label{fig:Clustersprocess2}%
\end{figure*}

Moreover, for the data-set that cannot ensure these conditions (see \figurename~\ref{fig:Clustersprocess2}), we use the second trained model that can combine detected clusters according to a threshold of common points. The appropriate threshold depends on the data-set structure and can be detected by brute force. \\



Therefore, the data-set is analyzed firstly to determine if clusters are overlapping or not. Based on the result, we select the appropriate trained model to detect its corresponding initialization parameters. \\




The unsupervised algorithms that are studied have different initialization parameters that should be configured carefully in the experiments. Thus, two versions of k-means were used. The first one is the naive k-means, that was tested with: the EM approach using random initial centroids, and the proposed solution. The second k-means implementation is the scikit-learn implementation, that was tested with: the k-means++ algorithm to choose the initial centroids, and the proposed solution.


\subsection{Clusters Number Correctness}

\begin{figure*}[!htbp]
\centering
\includegraphics[scale=0.35]{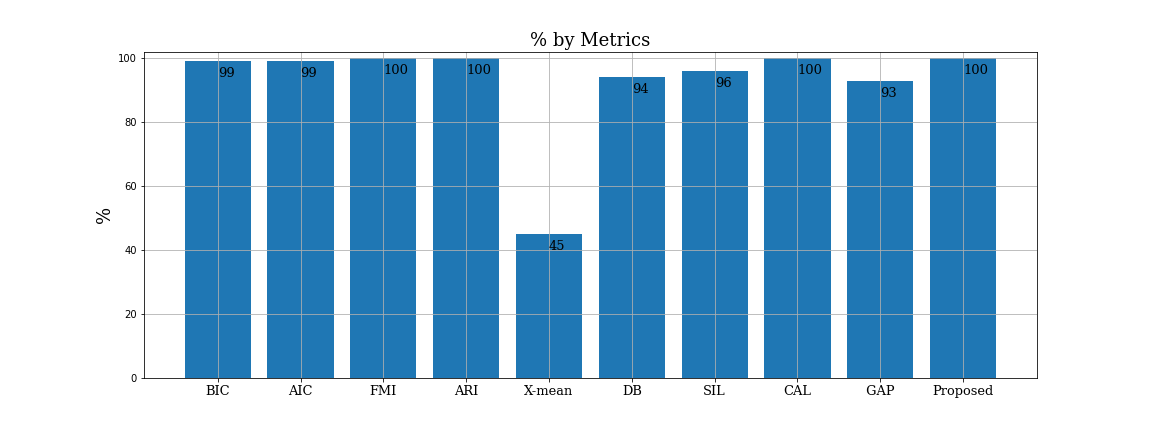}
\caption{Percentage of correctly detected number of clusters for 100 tests iterations, where clusters for each test iteration have been generated randomly with different clusters number, sizes and locations}
\label{fig:ClustersIdentified}%
\end{figure*}

In this part, we compare the performance of the existing clusters number estimation methods with our proposed solution. First, we use the YOLO-v5 trained model to detect the number of clusters. \figurename~\ref{fig:ClustersIdentified} presents the percentage of correctly predicted number of clusters for 100 test iterations, where clusters for each testing image (test iteration) have been generated randomly with different number of samples, sizes and locations. \\




The results indicate clearly that the proposed approach can detect correctly the clusters number. The clusters number detection rate with other validity metrics such as CH, SW, DB, Gap statistic, are shown in \figurename~\ref{fig:ClustersIdentified}. Certain metrics present high accuracy in detecting the number of clusters, similarly to our proposed solution. However, the advantage of our proposed solution compared to existing internal metrics is that the proposed approach requires less computation and memory overhead. Moreover, the proposed solution can work with partial data (a part of the whole clusters identified randomly, e.g. 20\% of samples). This leads to reduce more and more the required latency and resource overhead without degrading the accuracy of the clusters number estimation.


\subsection{Centroids Detection Correctness}
\begin{figure*}[!htbp]
\centering
\subfloat[][$k$=2]{\includegraphics[scale=0.125]{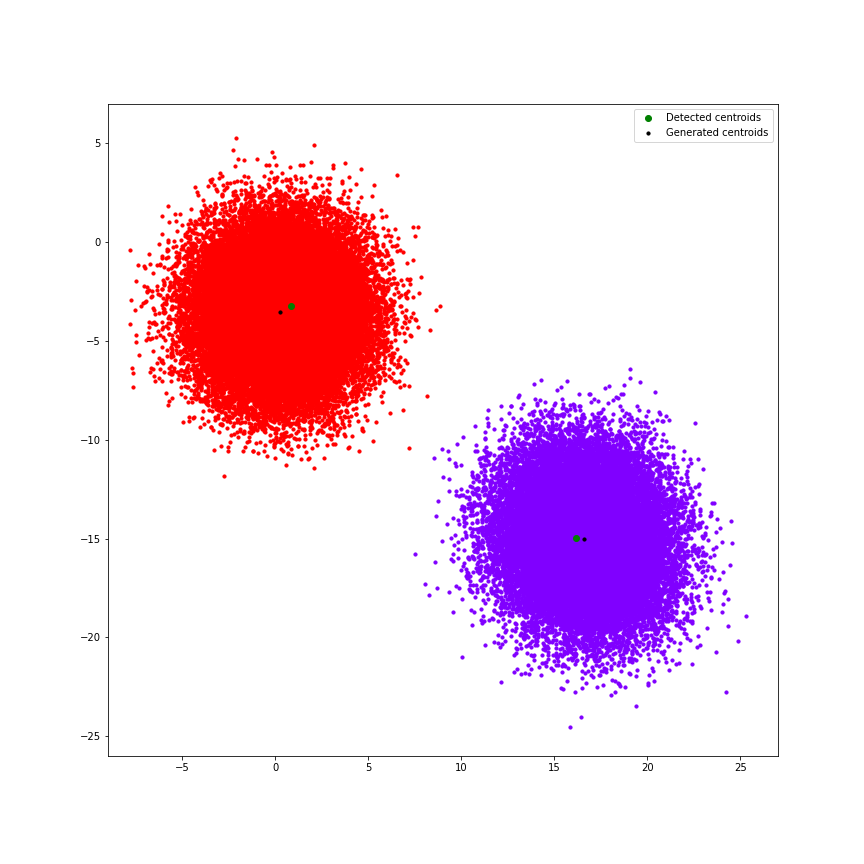}}
\subfloat[][$k$=3]{\includegraphics[scale=0.125]{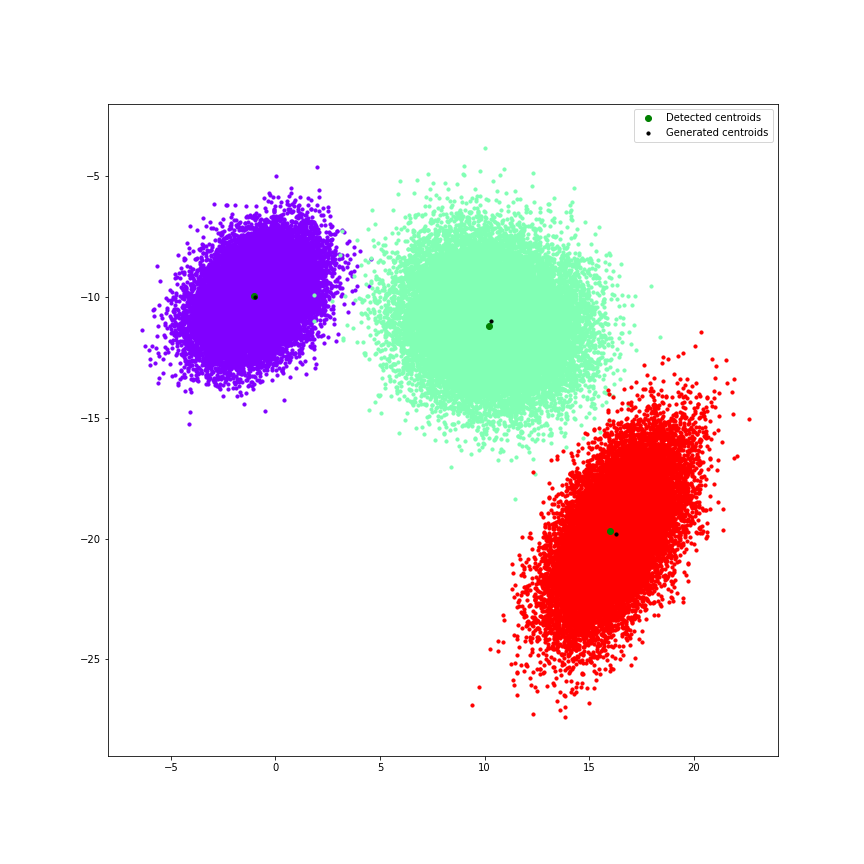}}
 \subfloat[][$k$=4]{\includegraphics[scale=0.125]{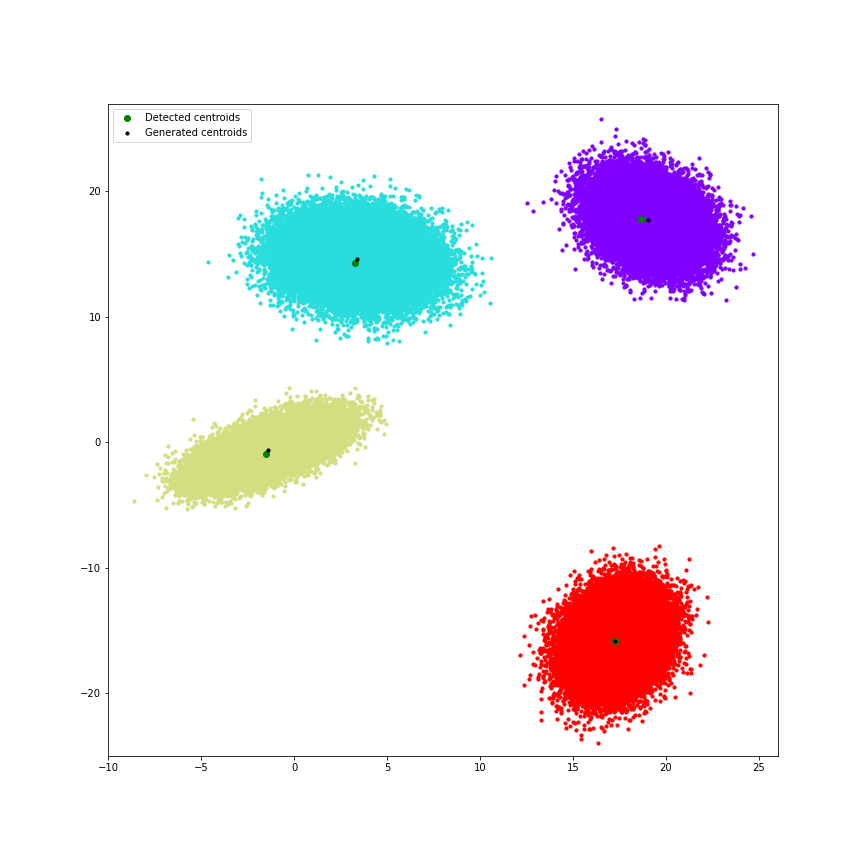}}
\subfloat[][$k$=5]{\includegraphics[scale=0.125]{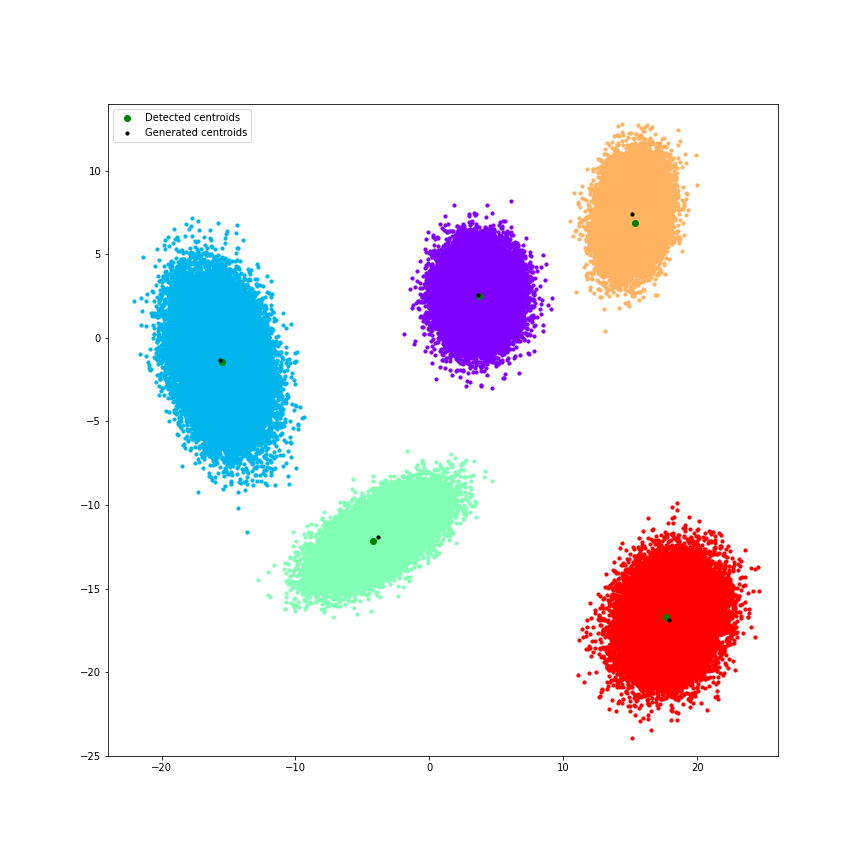}}\\\vspace{-0.42cm}
\subfloat[][$k$=6]{\includegraphics[scale=0.125]{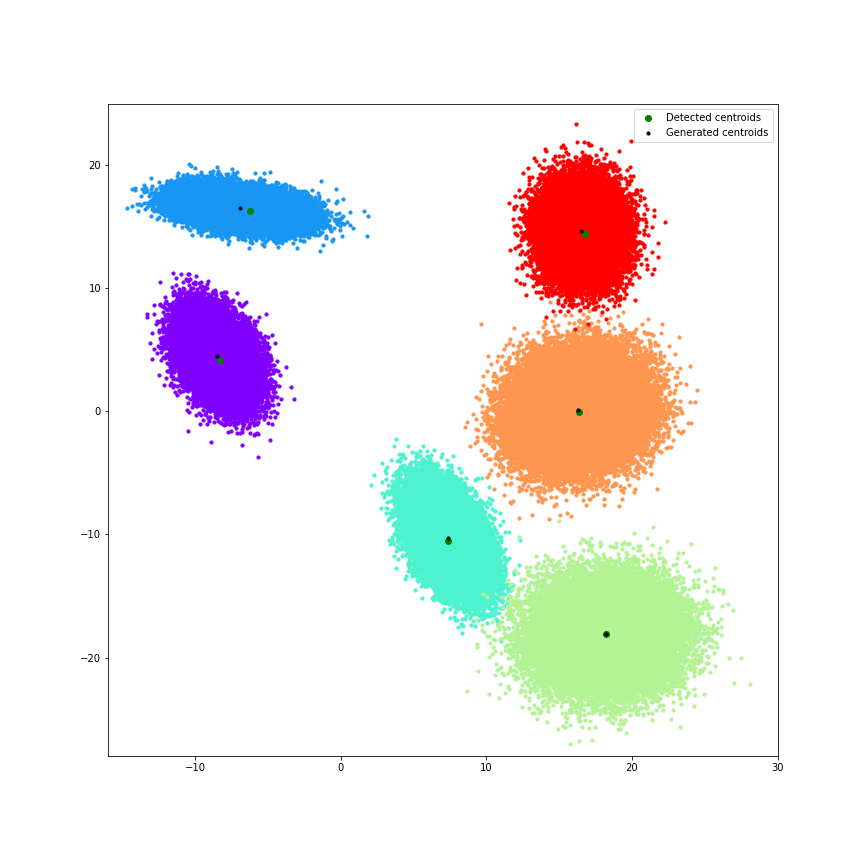}}
\subfloat[][$k$=7]{\includegraphics[scale=0.125]{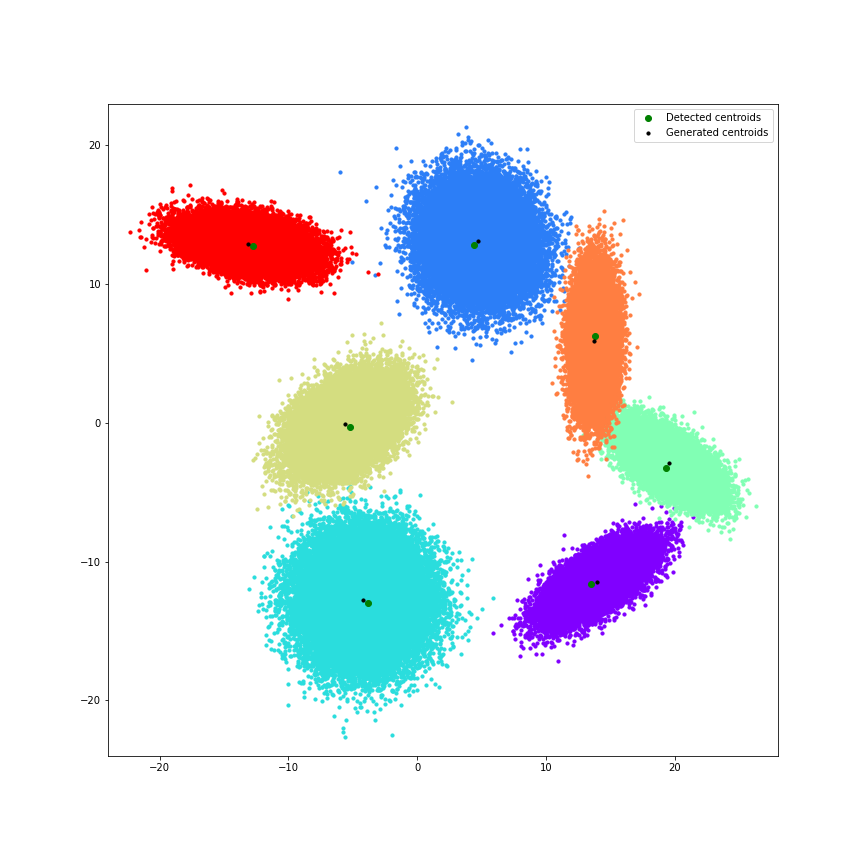}}
\subfloat[][$k$=8]{\includegraphics[scale=0.125]{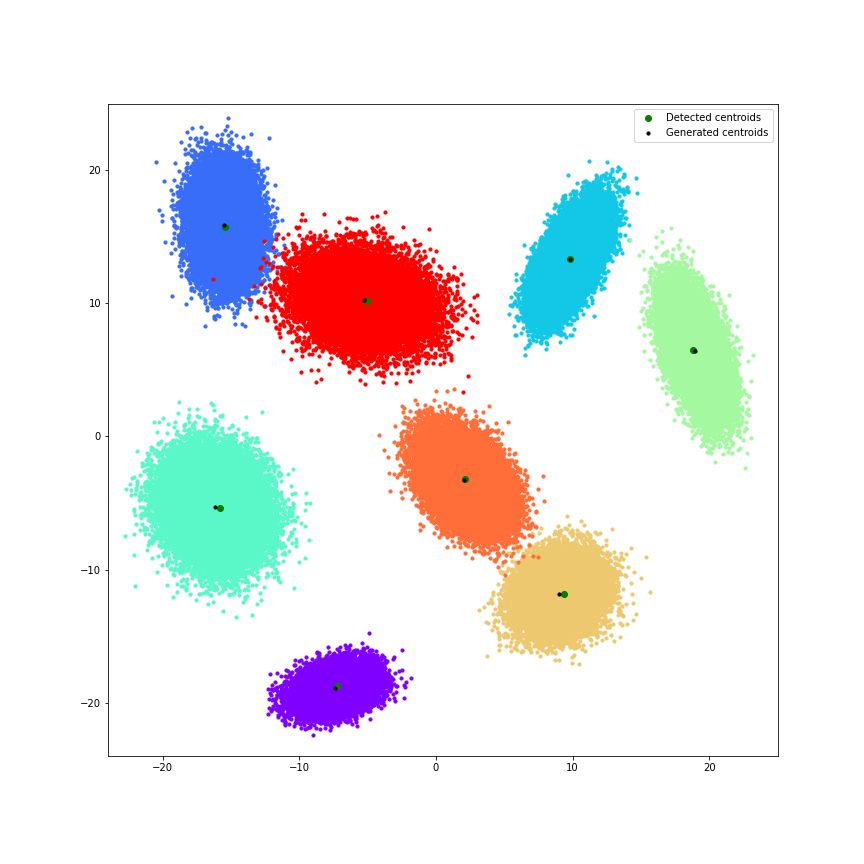}}
 \subfloat[][$k$=9]{\includegraphics[scale=0.125]{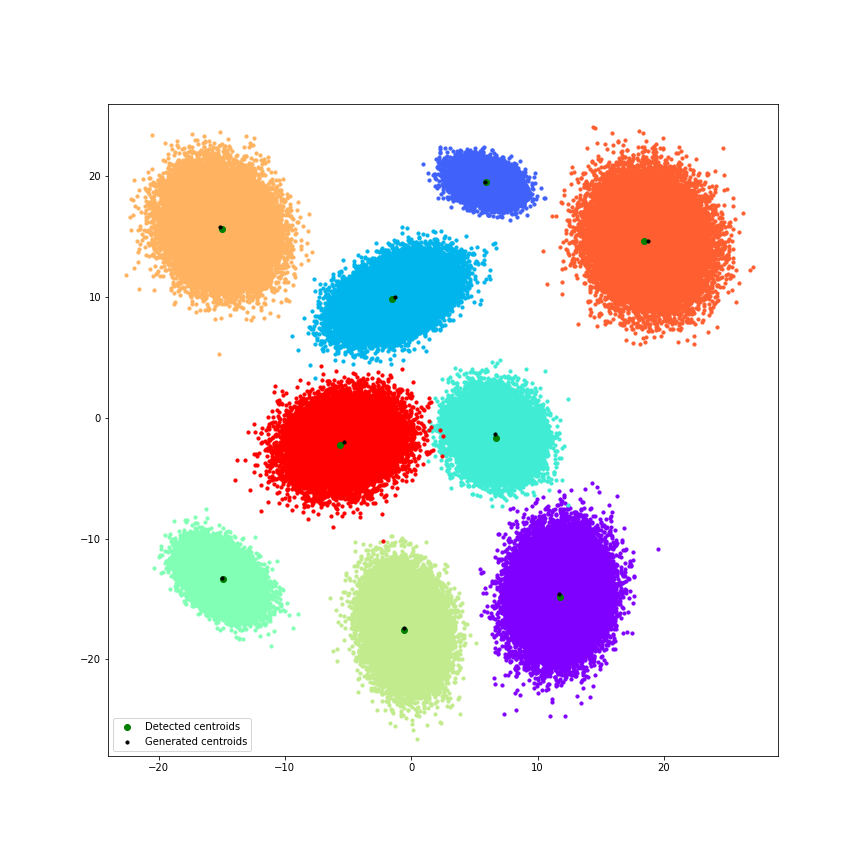}}\\
\caption{An example of validation data-sets with generated and identified centroids (consequently of each cluster)}%
\label{fig:ClustersValidation}%
\end{figure*}

\begin{figure*}[!htbp]
\centering
\includegraphics[scale=0.35]{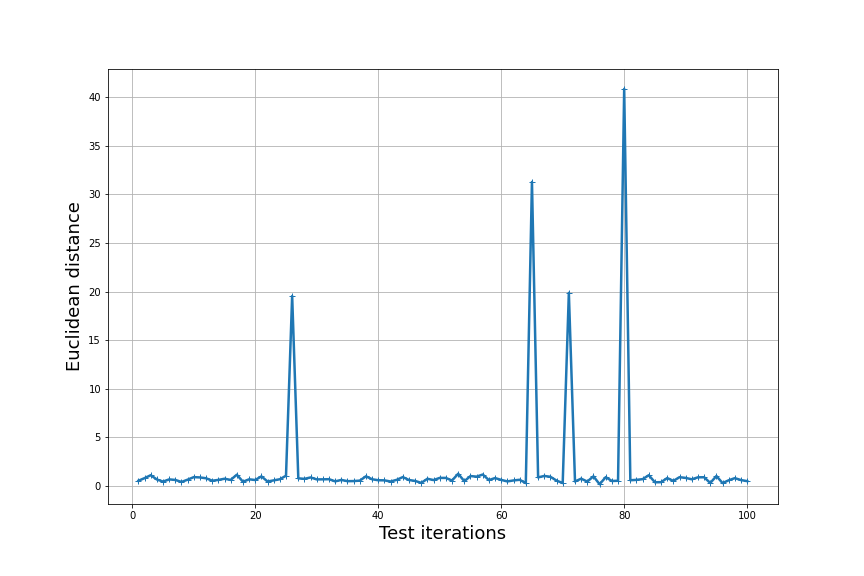}
\caption{Euclidean distance between generated and detected centroids for 100 test iterations: each test iteration consists of a set of clusters that have generated with random parameters such number of samples, size and location}
\label{fig:ClustersCentroids}%
\end{figure*}

In this part, we analyze the correctness of the detected centroids. \figurename~\ref{fig:ClustersValidation} presents a set of random generated testing images, where each one consists of a set of clusters produced using random configurations (centroids, sizes, and locations). We can remark visually that the identified centroids are very close to the generated ones for the different clusters (see \figurename~\ref{fig:ClustersValidation}). Moreover, we measure the euclidean distance between identified and generated centroids. The obtained results, presented in  \figurename~\ref{fig:ClustersCentroids}, indicate clearly that the generated and detected centroids are very close given that overall the obtained distances are less than one. Let us indicate that the euclidean distance between two points $p$ and $q$ in the $d$ space can be computed according to the following equation:  
\begin{equation}
   d\left( p,q\right)   = \sqrt {\sum _{i=1}^{d}  \left( q_{i}-p_{i}\right)^2 }   
\end{equation}
The lower is $d\left( p,q\right)$, the closer are the points $p$ and $q$. When $d$ is equal to 0, $p$ and $q$ are collocated. \\

These results show that the proposed method provides better approximation of the correct centroids and consequently this will result in increasing the accuracy of the clustering algorithm, in addition to reducing the memory consumption and the number of iterations for the clustering algorithms to converge. 

\subsection{Accuracy Rate}
In this part, the accuracy of the clustering algorithms is analyzed by using the proposed detected initialized parameters for k-means and other clustering algorithms.

The clustering accuracy rate (AR) can be computed as follows:
 \begin{equation}
 AR=\sum_{i=1}^k \frac{n(c_k)}{n}    
 \end{equation}
 where $n(c_k)$ is the number of data points that were correctly included in cluster $k$, and $n$ is the total number of data points. The higher is the $AR$, the better is the clustering correctness.\\

\begin{figure*}[!htbp]
\centering
\subfloat[][$nc$=2]{\includegraphics[scale=0.08]{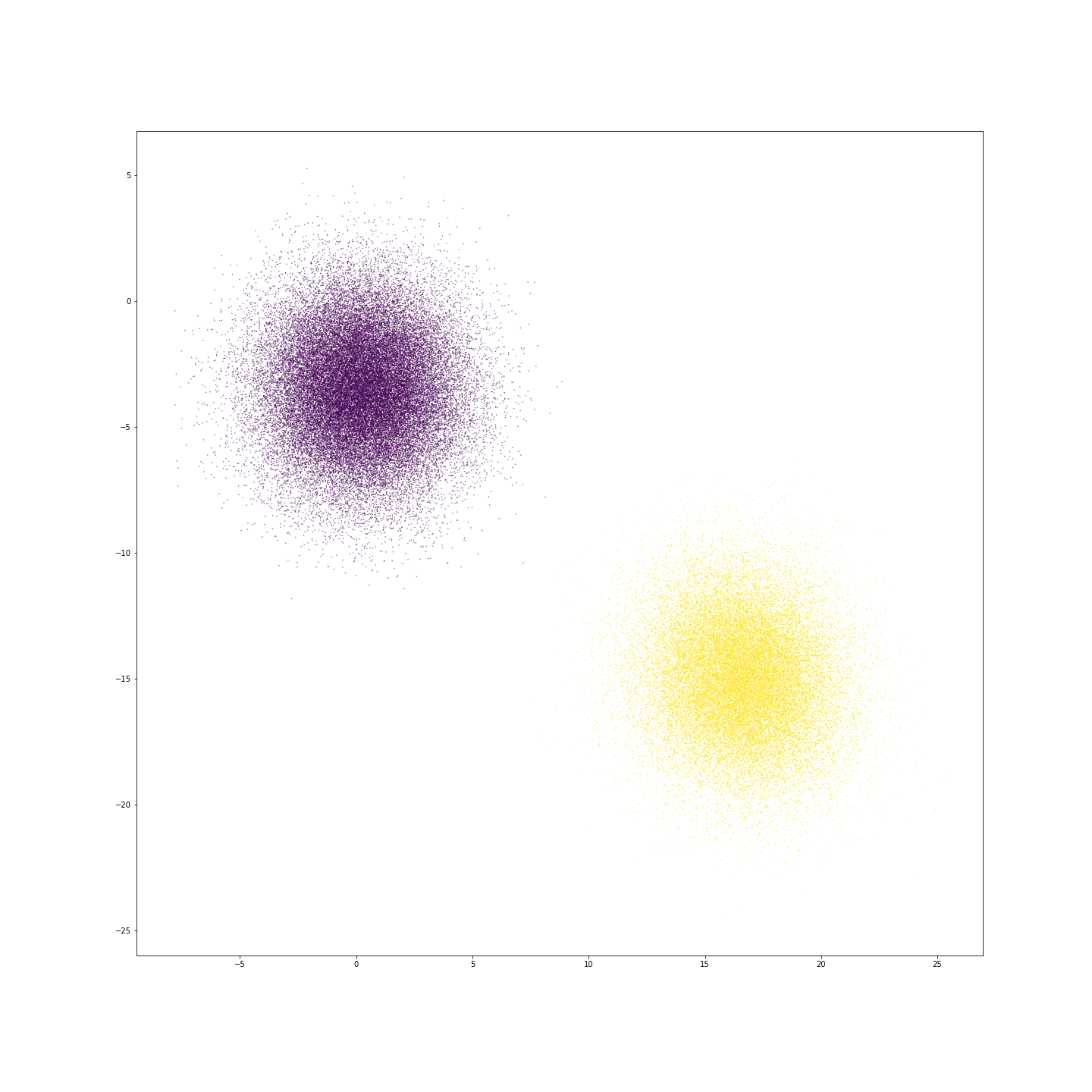}}
\subfloat[][$nc$=3]{\includegraphics[scale=0.08]{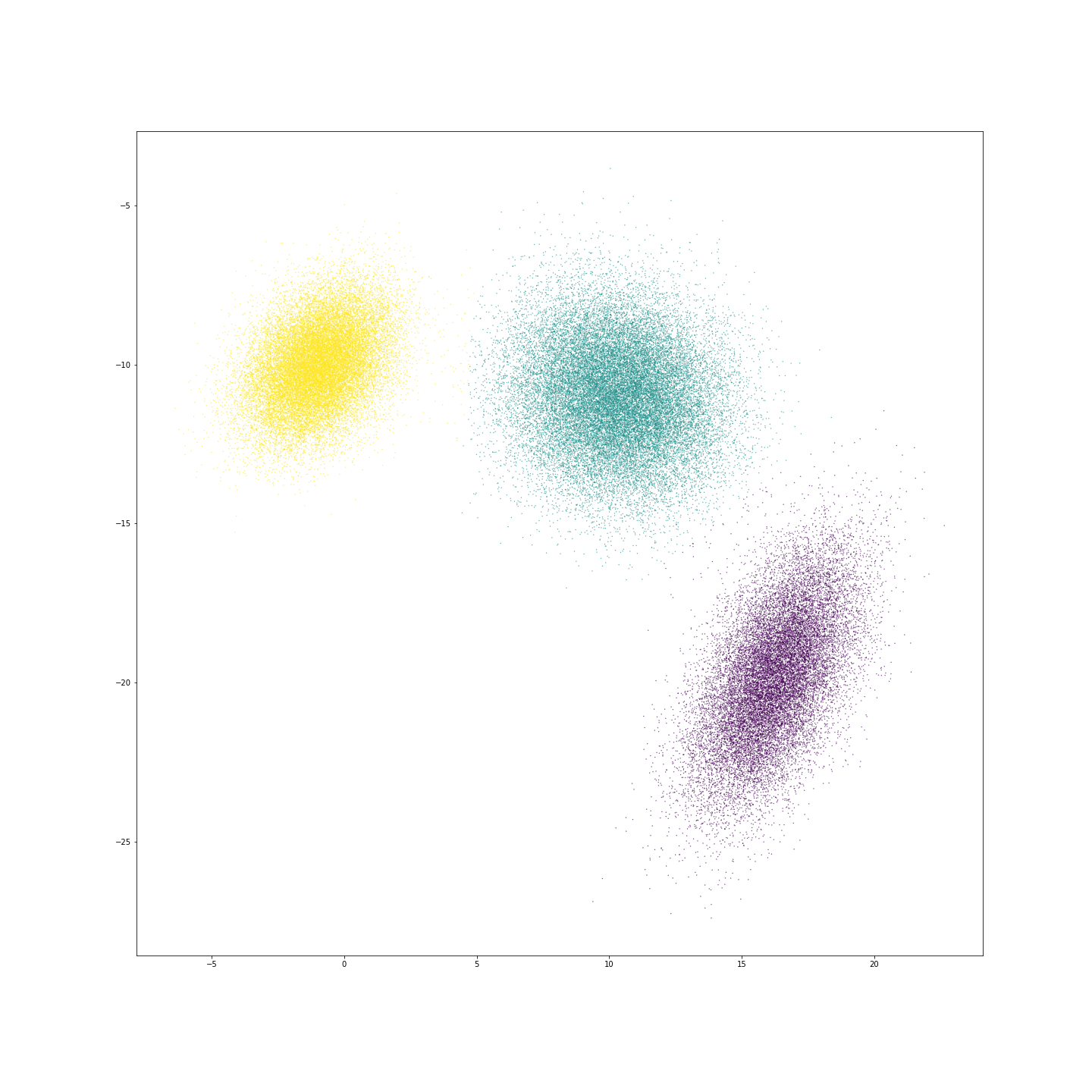}}
\subfloat[][$nc$=4]{\includegraphics[scale=0.08]{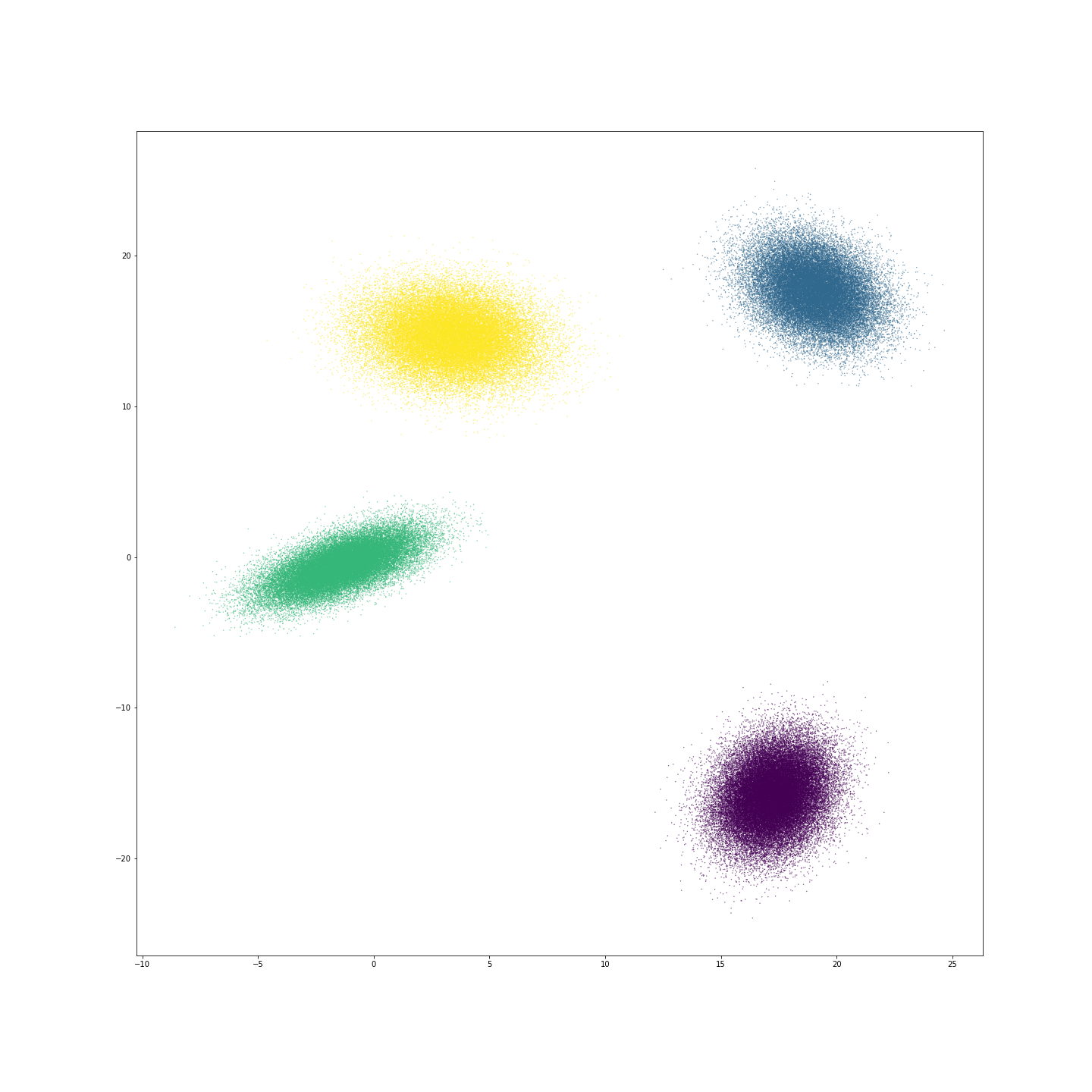}}
\subfloat[][$nc$=5]{\includegraphics[scale=0.08]{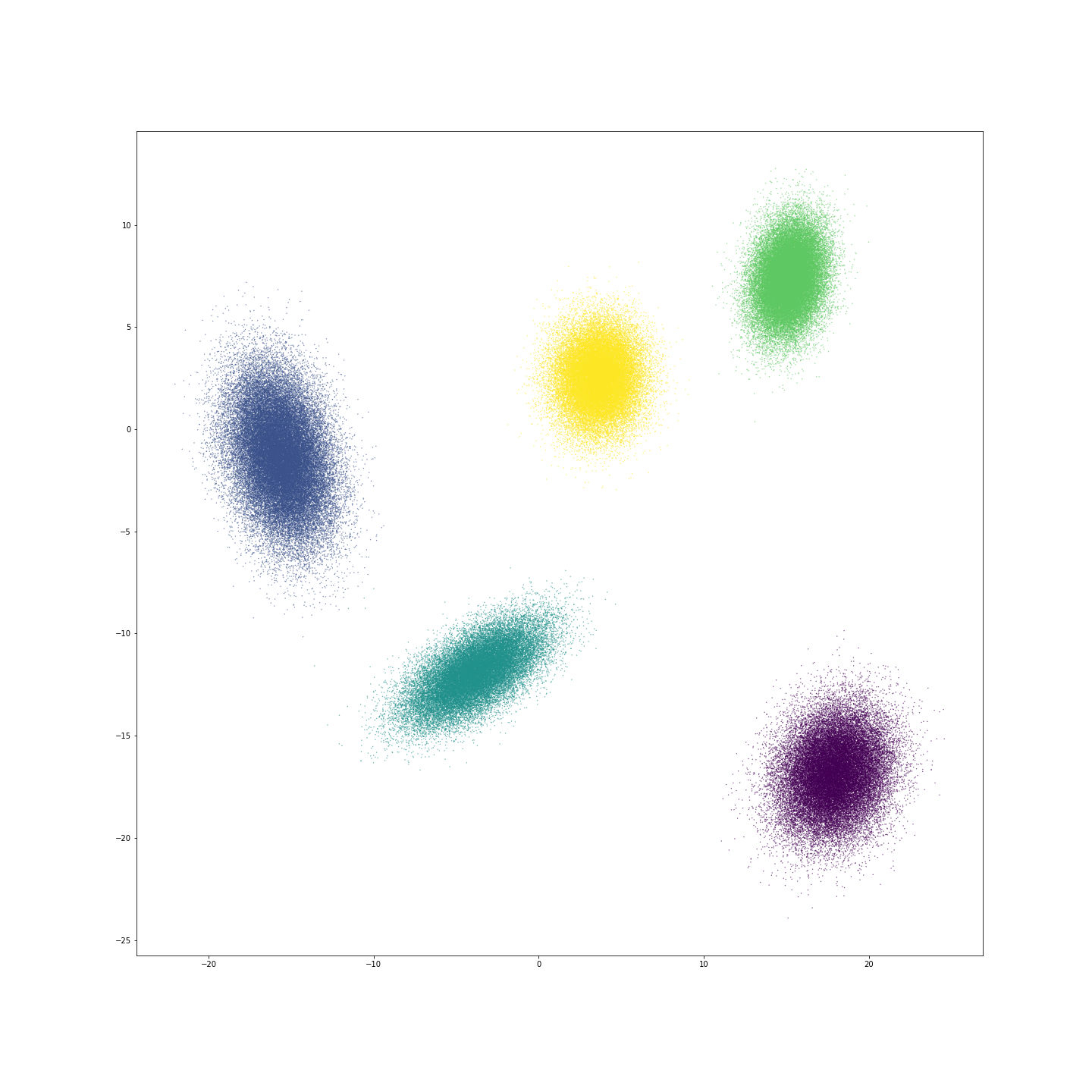}}\\
\subfloat[][$nc$=6]{\includegraphics[scale=0.08]{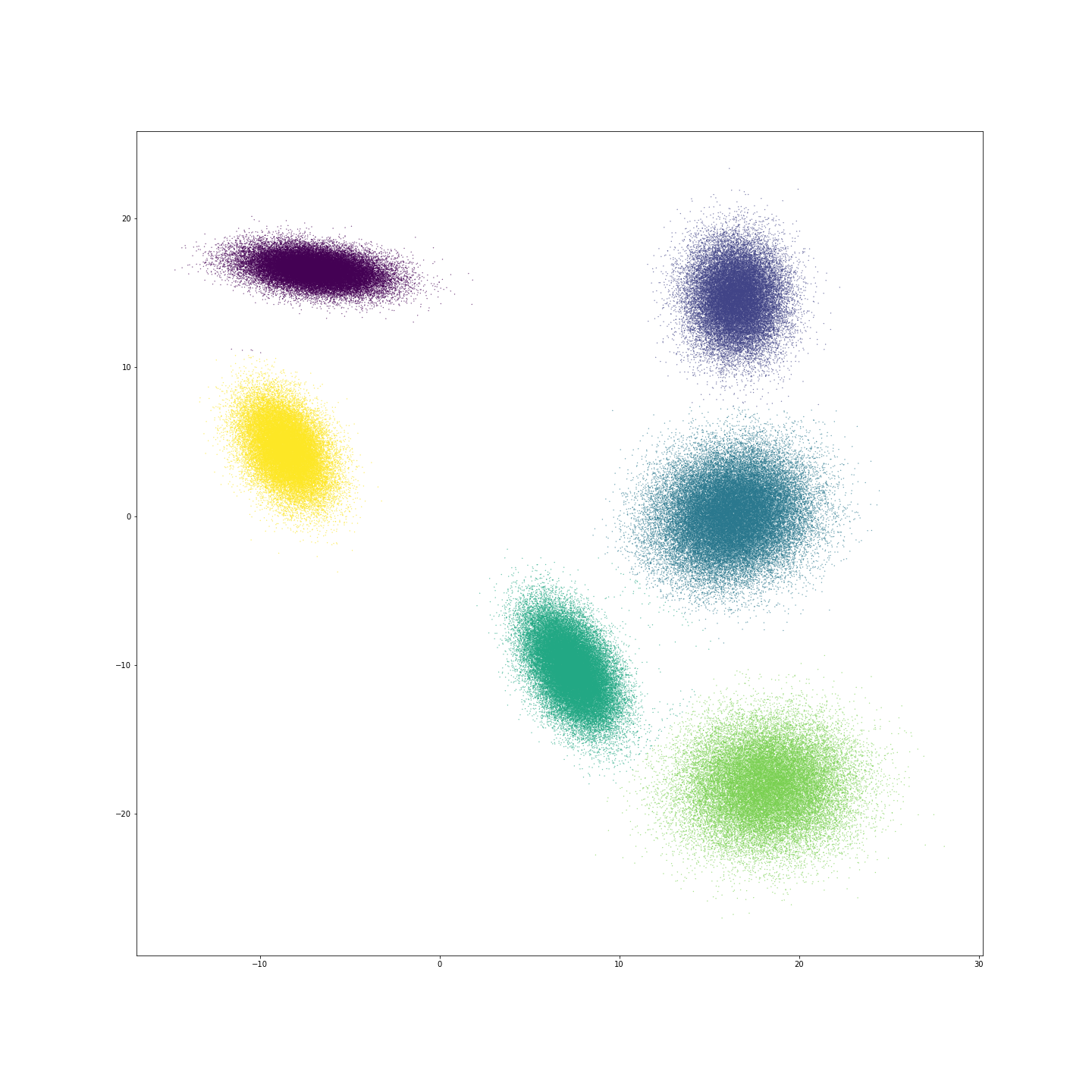}}
\subfloat[][$nc$=7]{\includegraphics[scale=0.08]{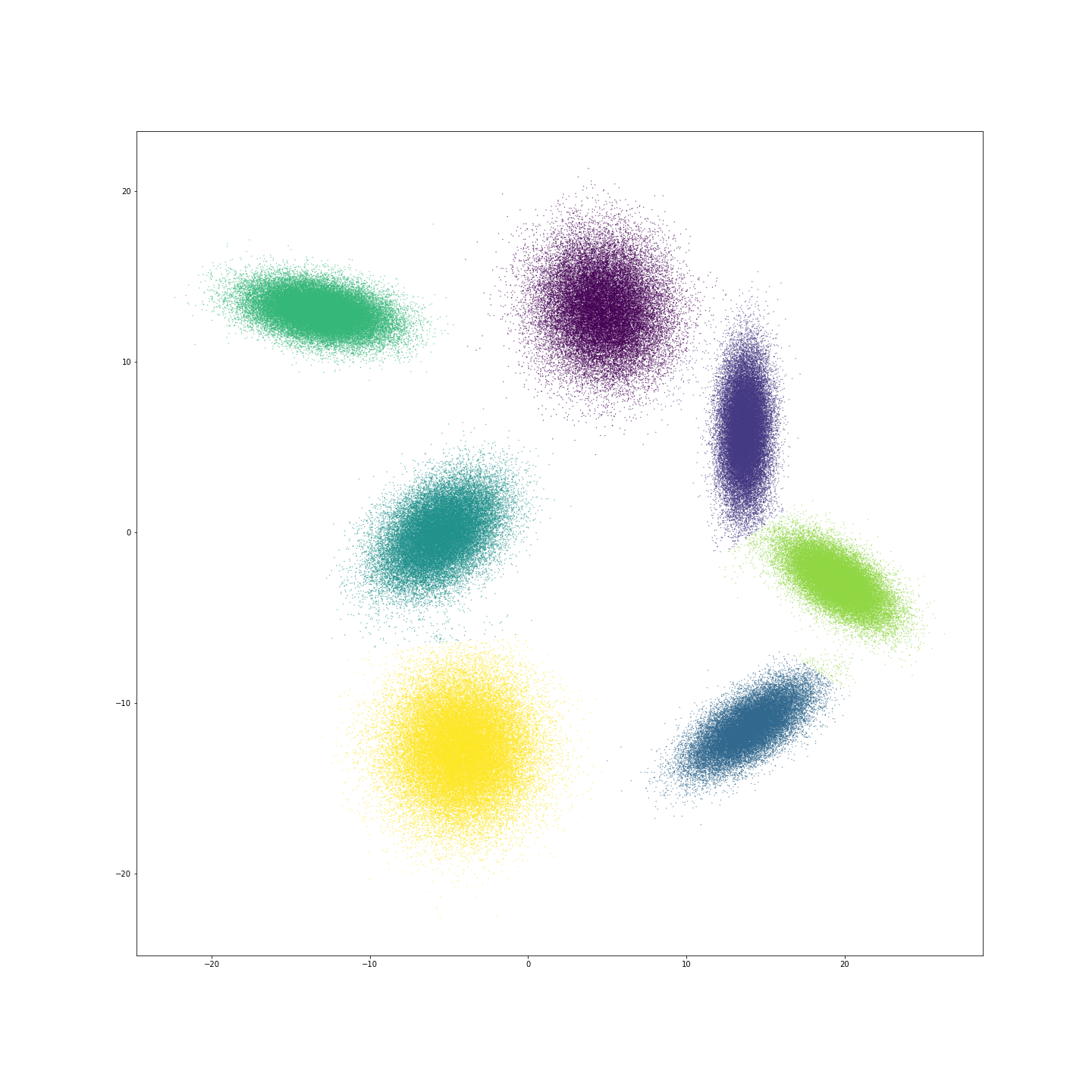}}
\subfloat[][$nc$=8]{\includegraphics[scale=0.08]{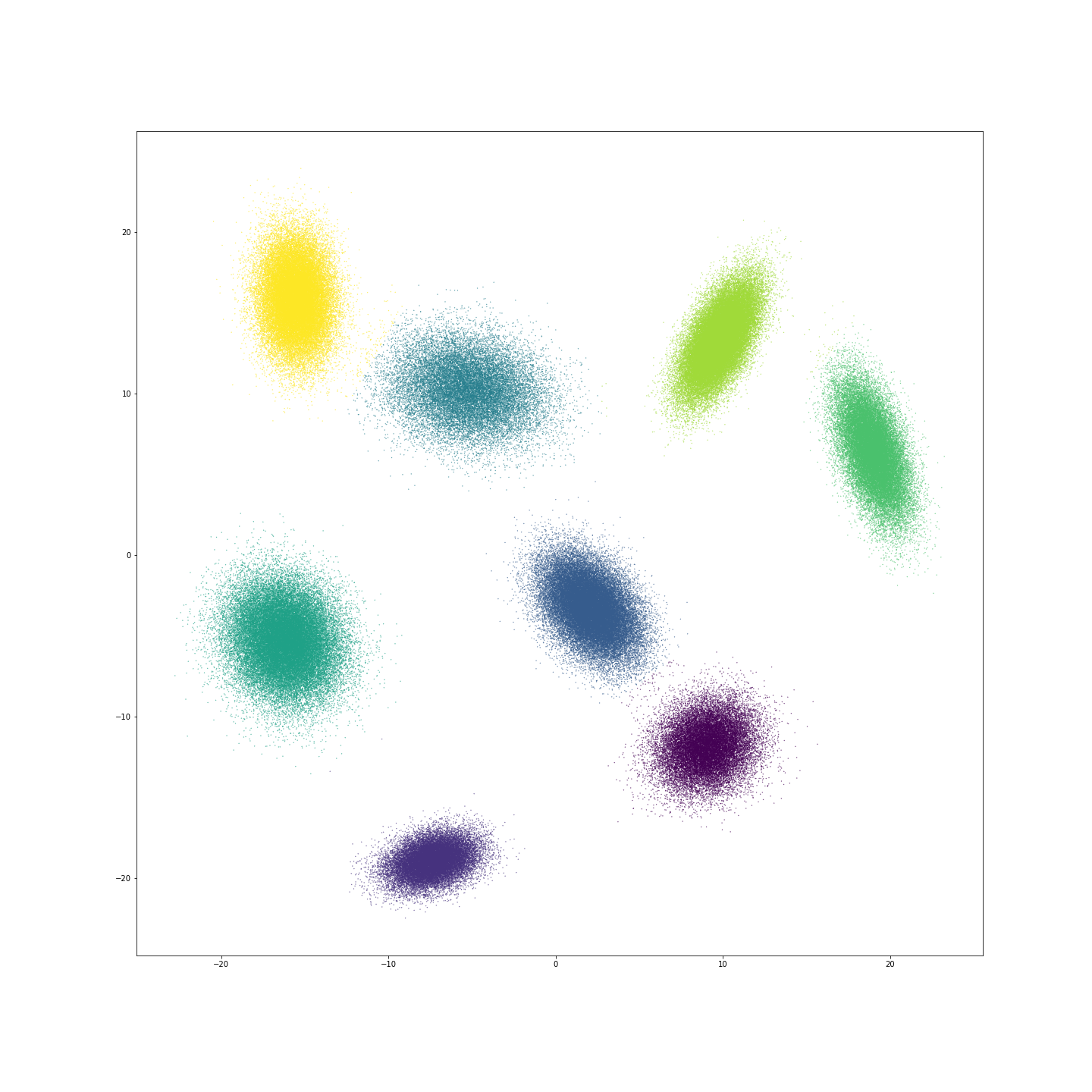}}
\subfloat[][$nc$=9]{\includegraphics[scale=0.08]{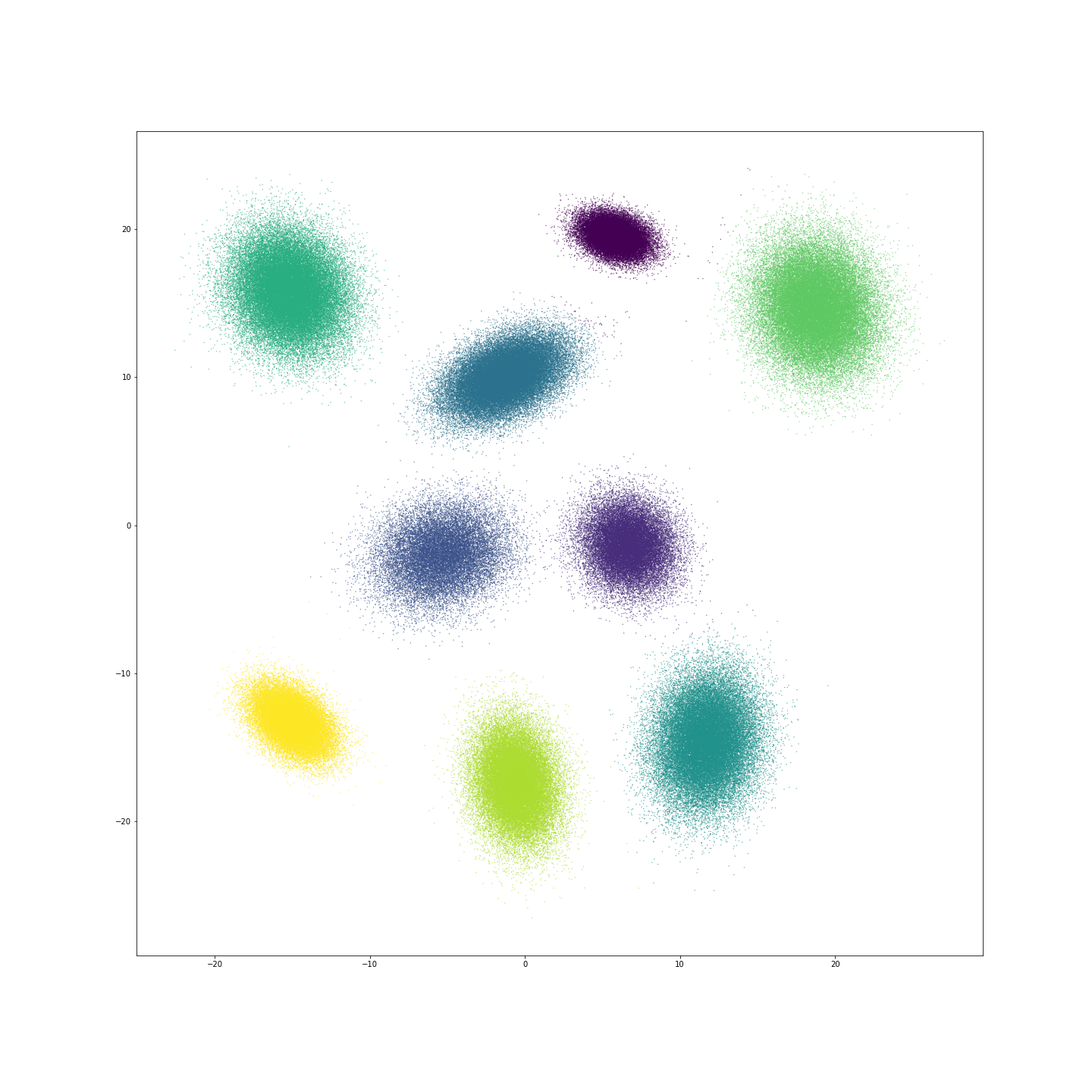}}\\\vspace{-0.22cm}
\caption{The k-means clustering results for the testing data-sets, listed in \figurename~\ref{fig:ClustersValidation},  by using the detected cluster number and identified centroids (proposed method) }%
\label{fig:Clustersprocess}%
\end{figure*}

\begin{figure*}[!htbp]
\centering
\subfloat[][Naive implementation]{\includegraphics[scale=0.32]{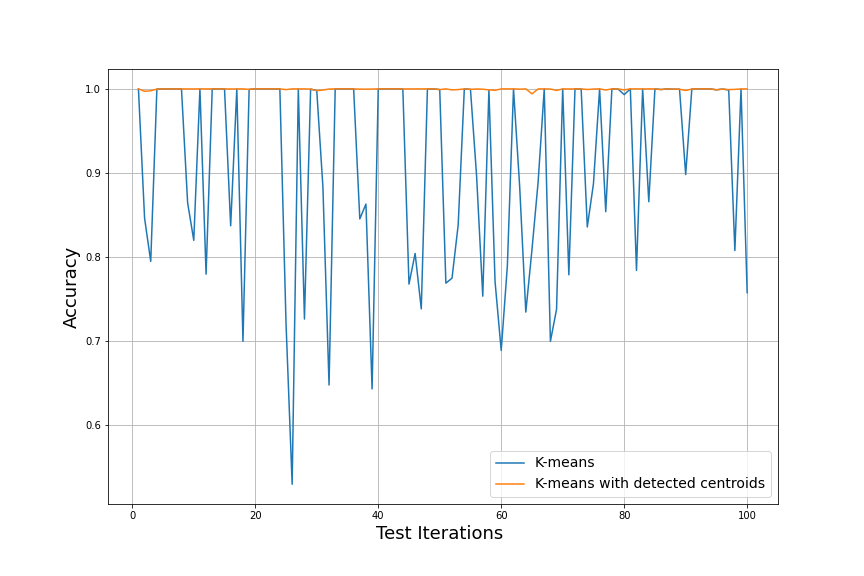}}
\subfloat[][Scikit-learn implementation]{\includegraphics[scale=0.32]{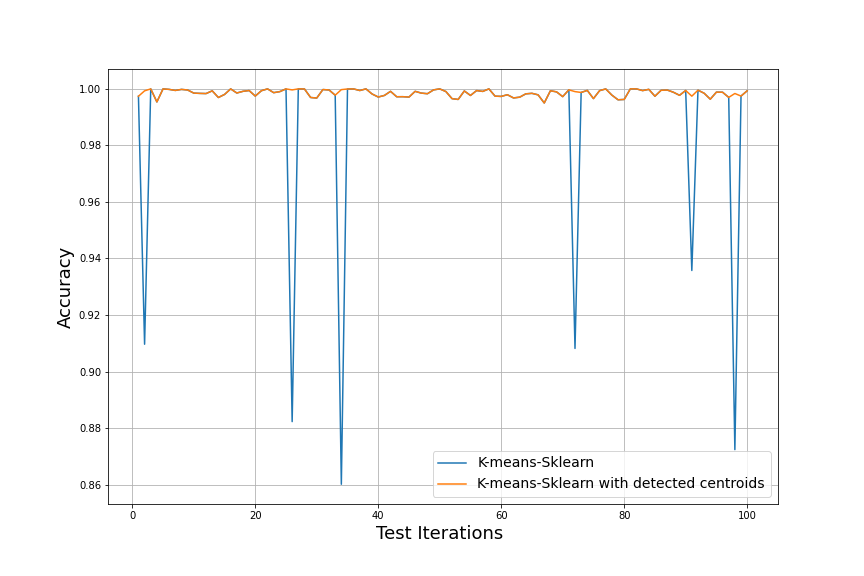}}
\caption{Accuracy of k-means clustering algorithms with and without using the proposed approach for two k-means implementations}%
\label{fig:AccuracyClustering}%
\end{figure*}

{\figurename~\ref{fig:AccuracyClustering} represents the $AR$ results of k-means with and without the identified centroids. These results validate that the clustering accuracy (at the samples level) is enhanced (to be close to 1) by using the identified centroids for a naive implementation of k-means and also for the optimized scikit-learn k-means implementation. Therefore, k-means (or any other clustering algorithm) can provide better clustering accuracy and low computational overhead (low number of iteration to converge and consequently delay) by using the proposed initialization parameters detection method. For example, as shown \figurename~\ref{fig:Clustersprocess}-e (detected $nc=$6), the accuracy is enhanced from 0.854 to 0.999 by using the identified clusters number and centroids.\\}


\begin{figure*}[!ht]
\centering
\subfloat[][Naive K-means implementation]{\includegraphics[scale=0.2]{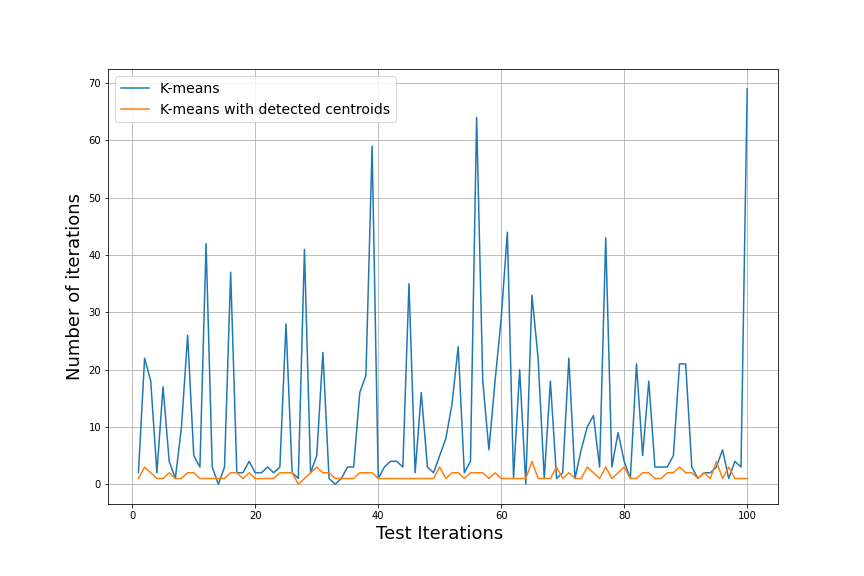}}
\subfloat[][Scikit-learn K-means implementation]{\includegraphics[scale=0.2]{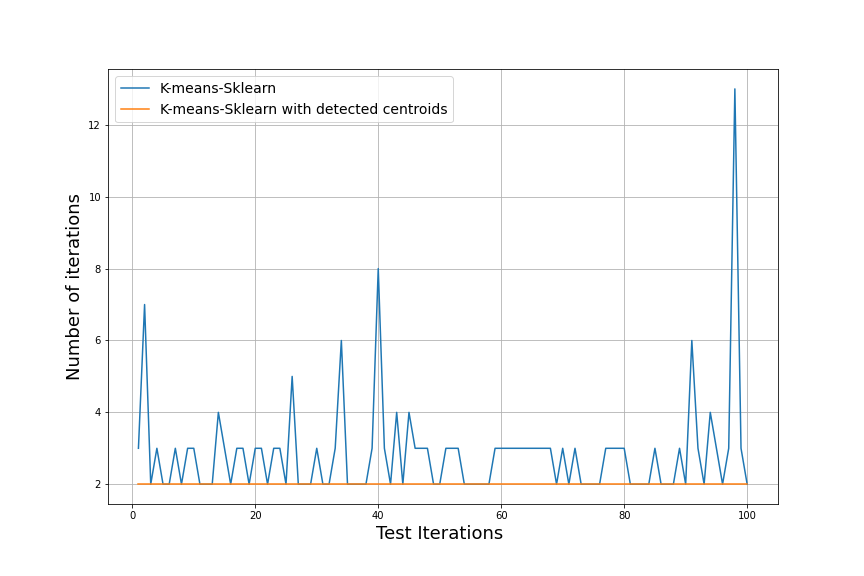}}
\subfloat[][FCM]{\includegraphics[scale=0.2]{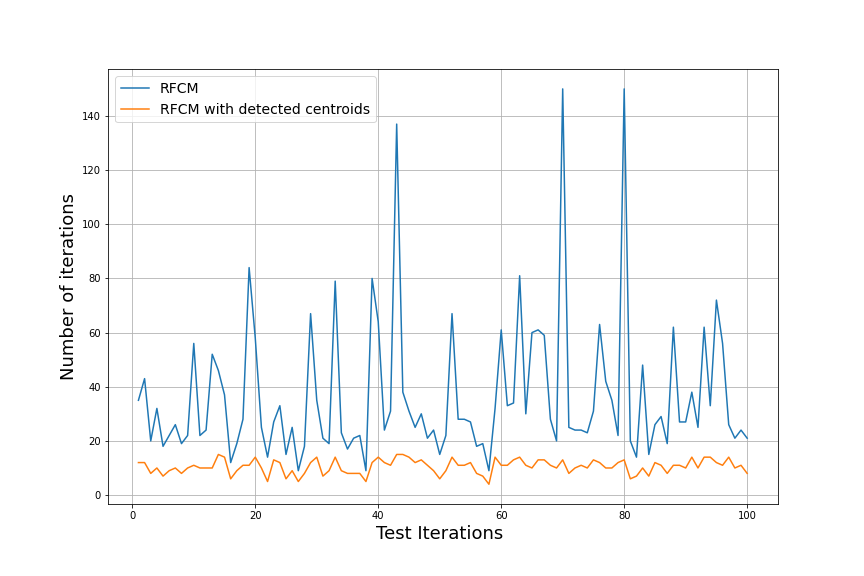}}
\caption{Variation of number iterations required to converge for k-means(a), optimized implementation with scikit-learn(b) and for FCM(c) in function of test iterations}%
\label{fig:Iterations}%
\end{figure*}

On the other hand, in case of $nc=7$ (\figurename~\ref{fig:Clustersprocess}-f), the required number of iterations to converge is decreased from 4 to 2 by using the identified centroids. Moreover, \figurename~\ref{fig:Iterations}  confirms this reduction since the required number of iterations to converge is reduced for k-means and FCM when using the proposed approach compared to the traditional clustering approaches (for 100 test images).


\subsection{Execution Time}
In general, the time delay of any clustering algorithm depends on three factors:
\begin{enumerate}
    \item The number of the algorithm iterations
    \item The amount of data points 
    \item The time needed to find out the clustering centers and data points partitions
\end{enumerate}

\begin{figure*}[!ht]
\centering
\includegraphics[scale=0.35]{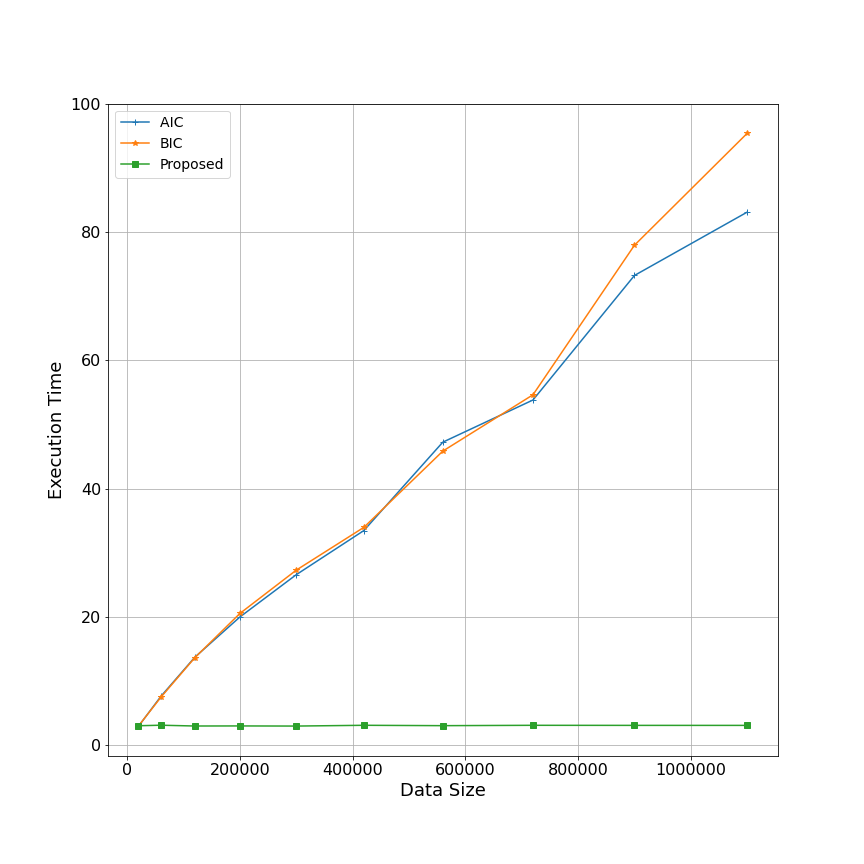}
\caption{Variation of execution time (in seconds) to detect correct number of clusters by using the proposed approach, AIC, and BIC}
\label{fig:ClusterstimeIdentification}%
\end{figure*}

First of all, we computed the testing time of the proposed solution to detect cluster initialization parameters. \figurename~\ref{fig:ClusterstimeIdentification} shows the execution times (in seconds) of two well-known internal indices methods (AIC and BIC) and the proposed method versus data sizes. We observe lower execution time for the proposed solution compared to the existing indices methods, making it the best choice when working on large data-sets. The proposed solution has the advantage of requiring very low computation complexity and consequently testing delay since it is independent of the number of data points. However, with the increase in the data-set size, the running time for all internal metrics is significantly increased. \\

\begin{figure*}[!ht]
\centering
\includegraphics[scale=0.35]{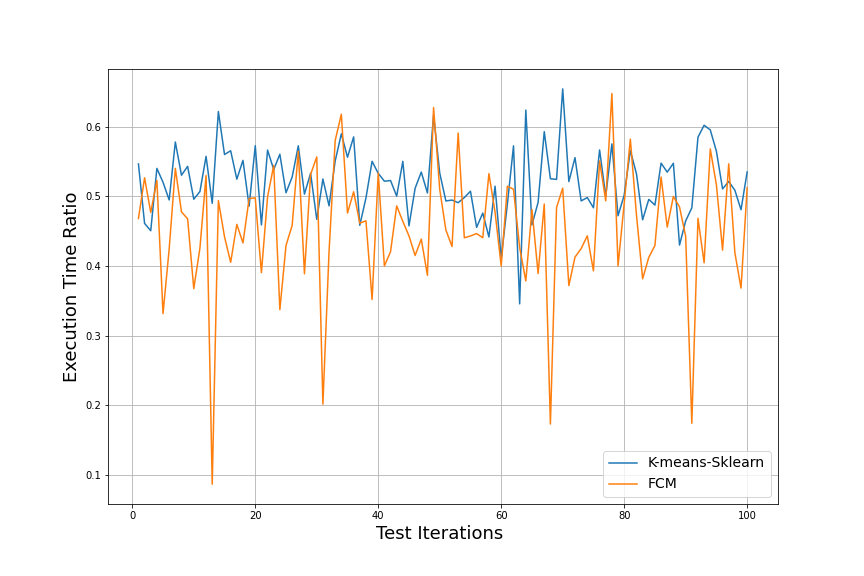}
\caption{Ratio of the execution time for k-means and FCM between with and without the detected clusters initialization }
\label{fig:Clusterstime}%
\end{figure*}

Moreover, we quantify the effectiveness of using these initialization parameters to help clustering algorithm converging fast. \figurename~\ref{fig:Clusterstime} shows the execution time ratio between using clustering algorithms with identified centroids and without. These results show clearly that using the identified centroids reduces the testing time to half in average. This indicates that the proposed solution reduces the clustering testing time and makes it suitable for real time applications.

\subsection{Flexibility}
Another important property, that the proposed solution can provide, is the flexibility to deal with different clusters forms. The proposed solution can be extended by adding existing cluster formats such as: blobs, blobs with varied variances, noisy circles, noisy moons, no structure, and Anisotropicly distributed data, which are described and presented in~\cite{www:scikit-learn.org}.

\section{Conclusion}
\label{sec:conc}

This paper proposes a new clustering initialization method that can determine the number of clusters in addition to their possible centroids and sizes. The proposed solution uses a DL-based object detection model (YOLO-v5) to detect the initial clustering parameters. The advantages of the proposed solution is that it is: lightweight, fast, and robust with different cluster volumes, shapes, and noise. The proposed solution has been realized with several configurations, demonstrating its efficiency compared to existing approaches, especially in terms of time complexity and resources overhead. Two models are presented. The first one is for the separated clusters and the second one is for overlapping clusters. The choice of which model to use is determined based on the characteristics of the input dataset. As future work, we aim to design a new DL-based data transformation model towards getting well-separated clusters. As well, we aim to work with 3D object detection models to be applied on 3D clusters representations, in addition to consider other cluster forms.


\section*{Acknowledgment}
This work was  partially  supported  by  the  EIPHI  Graduate  School(contract ”ANR-17-EURE-0002”). The Mesocentre of Franche-Comté provided the computing facilities.

\end{document}